\documentclass[lettersize,journal]{IEEEtran}  

\usepackage[hyphens]{url}  
\usepackage{algorithm}
\usepackage{algorithmic}

\usepackage{amsmath,amssymb,amsfonts}
\usepackage{makecell}
\usepackage{subfigure} 
\usepackage[english]{babel}

\usepackage{textcomp}
\usepackage{stfloats}
\usepackage{url}
\usepackage{verbatim}
\usepackage{graphicx}
\usepackage{newfloat}
\usepackage{listings}
\usepackage{hyperref}
\usepackage[T1]{fontenc}
\usepackage{aecompl}
\usepackage{booktabs}
\usepackage{multirow} 
\usepackage{color,xcolor}
\usepackage{threeparttable} 
\usepackage{cleveref}
\usepackage{cite}
\def\BibTeX{{\rm B\kern-.05em{\sc i\kern-.025em b}\kern-.08em
    T\kern-.1667em\lower.7ex\hbox{E}\kern-.125emX}}
\usepackage{balance}
\begin{document}
\title{Reinforcement Learning-Based Pathfinding for Multiple UAVs Facing Abrupt Hazardous Areas}

\author{Qizhen Wu, Lei Chen, Kexin Liu, \IEEEmembership{Member, IEEE}, and Jinhu L\"u, \IEEEmembership{Fellow, IEEE}
\thanks{*This work was supported in part by the National Key Research and Development Program of China under Grant 2022YFB3305600, and in part by the National Natural Science Foundation of China under Grants 62141604, 62088101, and 62003015. ({\it{Corresponding
author: Lei Chen.}})}
\thanks{Qizhen Wu, Kexin Liu, and Jinhu L\"u are with the School of Automation Science and Electrical Engineering,
        Beihang University, Beijing 100191, China. (e-mail: wuqzh7@buaa.edu.cn; skxliu@163.com; jhlu@iss.ac.cn)}%
\thanks{Lei Chen is with the Advanced Research Institute of Multidisciplinary Sciences and State Key Laboratory of CNS/ATM, Beijing Institute of Technology, Beijing 100081, China. (e-mail: bit$\_$chen@bit.edu.cn)}%
\thanks{The experiment video is available at \url{https://www.bilibili.com/video/BV1gw41197hV/?vd_source=9de61aecdd9fb684e546d032ef7fe7bf}.}
\thanks{The code is available at \url{https://github.com/Wu-duanduan/Pathfinding_MARL}.}
}

\maketitle

\begin{abstract}
     Planning feasible paths for multiple unmanned aerial vehicles (UAVs) amidst abrupt hazardous areas is a critical safety challenge, where existing methods often lack safety guarantees and uncertainty handling. To address this, we propose a novel multi-agent reinforcement learning (MARL) approach for the UAV pathfinding problem. Our method ensures rapid responsiveness and adherence to safety constraints through the integration of a control barrier function, guaranteeing safe replanning even during sudden route changes. To overcome the potential inefficiency of purely reactive safety, we introduce a probabilistic neural network that quantifies hazard uncertainty, enhancing the anticipation of sudden dangers. Finally, to utilize swarm intelligence for mutual risk avoidance, the approach incorporates neighbors' observations using a proximity-weighted mean-field mechanism, allowing each UAV to consider the impact of this aggregated information in its planning. Extensive simulations show that our method achieves a planning success rate surpassing 90\% in transient environments, outperforming traditional planners and other MARL baselines. Real-world experiments further validate the approach's adaptability, demonstrating its practical value for safety-critical missions.
\end{abstract}

\def\abstractname{Note to Practitioners}
\begin{abstract}
This paper is motivated by the safety-critical challenge of navigating UAV swarms through environments with abrupt, unpredictable hazards. Traditional planners often suffer from high computational latency during sudden changes, while standard learning-based methods typically lack rigorous safety guarantees. To address this, we present a decentralized pathfinding framework that integrates probabilistic uncertainty handling with a control barrier function. For implementation, the proposed system is lightweight enough for standard onboard embedded computers, enabling autonomous execution without centralized control. Experimental results indicate that the average computation time per step is less than 5 ms, leaving ample computational resources for other onboard tasks like state estimation and visual processing. This approach is particularly applicable to time-sensitive missions like disaster response, offering a solution that remains robust even under conditions of communication instability or sensor noise.
\end{abstract}

\section{Introduction}
\IEEEPARstart{W}{ith} rapid advancements in artificial intelligence \cite{wu2024hierarchical,fu2023robust}, unmanned aerial vehicles (UAVs) \cite{betalo2025dynamic,zhang2024efficient} have attracted considerable interest due to their efficiency, flexibility, and cross-domain potential. A critical challenge for UAVs is pathfinding, particularly in stochastic environments, requiring collision-free paths in the presence of dynamic obstacles and uncertain conditions \cite{ze2023time}. Abrupt hazardous areas, such as those caused by explosions, adverse weather, or unforeseen obstacles \cite{liu2019collision}, complicate pathfinding, necessitating adaptive real-time planning for safety. Traditional pathfinding methods, such as graph search \cite{guo2023hpo} and heuristic algorithms \cite{yan2024efficient}, perform well in static environments but struggle in dynamic and uncertain environments. These approaches often demand substantial computational resources for path replanning when disruptions occur \cite{heuer2023proactive}. Their limitations underscore the need for scalable solutions to address uncertainties, particularly for safety-critical UAV applications.

Recently, reinforcement learning (RL) methods for flight control have advanced significantly. UAVs trained with RL now outperform many human-operated platforms \cite{kaufmann2023champion,wang2025unlocking}. After extensive training, UAVs can control themselves precisely and rapidly in known environments. However, flight control and pathfinding are coupled challenges for UAVs. Planning feasible paths for UAVs in complex environments, especially for multiple UAVs in dynamic scenarios, remains challenging despite significant progress in flight control by researchers \cite{wang2024constrained,chowdhury2024interchangeable} and industry.

Multi-agent reinforcement learning (MARL) is an effective approach for enabling pathfinding in multiple UAV systems. Through iterative training with a shared objective \cite{wu2024hierarchical}, MARL integrates discrete task allocation and continuous path planning, enabling UAV collaboration. Furthermore, advanced MARL techniques \cite{zhao2024reinforcement,chen2022joint} enable UAVs to coordinate paths autonomously in partially observable environments with limited communication. Typically, MARL frameworks adopt one of three paradigms: centralized, decentralized, or centralized training with decentralized execution (CTDE). While centralized methods face scalability challenges and decentralized approaches can struggle with stability, CTDE offers a compelling balance. It allows UAVs to use shared data during centralized training to learn effective coordination, and plan optimal paths efficiently during decentralized execution. For example, the methods in \cite{zheng2024multi,yan2022} use CTDE to address UAV pathfinding for multi-task operations.

Despite its advantages, CTDE-based MARL methods primarily focus on path planning in static environments, where uncertainties are rarely addressed. Nevertheless, UAVs in real-world scenarios typically operate in environments with dynamic changes. The solutions proposed by these methods struggle to address dynamic uncertainties, leading to reduced robustness. A well-designed reward enables UAVs using RL to avoid periodically moving obstacles \cite{xu2025navrl}, but it cannot handle high-uncertainty scenarios. Recent work \cite{han2023multi, cao2023safe} has utilized the control barrier function (CBF) to enhance safety in RL. These approaches typically define a safe set via a barrier function and enforce constraints on the control input to guarantee the system state remains within this set. Beyond immediate safety, addressing the inherent stochasticity of the environment is vital. According to \cite{hu2023uncertainty}, quantifying uncertainty is critical for managing risks associated with abrupt hazards. It requires models that can enhance the anticipation of sudden dangers and reduce reliance on purely reactive mechanisms. Finally, in multi-agent scenarios, effective coordination is key for mutual risk avoidance. While shared observations among agents can improve situational awareness, unweighted mean-field MARL approaches \cite{wang20223m} that simply average neighbor information are insufficient. This uniform weighting can dilute critical warnings from nearby agents with less relevant data from distant ones, potentially obscuring immediate threats.

In general, the aforementioned studies primarily focus on environments with limited dynamics or lack formal safety guarantees for multiple UAVs. Here, we bridge this gap by proposing a novel MARL approach that enables UAVs to respond swiftly and safely to hazardous areas. With our approach, UAVs can replan their paths on short notice when facing hazards ahead. Specifically, we establish a foundation for safe flight during rapid route changes by integrating a CBF, which acts as a real-time safety filter. To enhance hazard anticipation beyond immediate reactions, we quantify the uncertainties caused by abrupt hazardous areas with a probabilistic neural network (PNN), which outputs a Gaussian distribution to capture potential environment changes. Additionally, to leverage swarm intelligence for mutual risk avoidance, the approach incorporates neighbors' observations using a proximity-weighted mechanism, allowing each UAV to consider the impact of this aggregated information in its planning. Extensive simulations and real-world experiments validate our approach, demonstrating planning success rates exceeding 90\% compared with baseline methods.
The key contributions of this work are:
\begin{enumerate}
\item To ensure rapid responsiveness facing sudden hazards, our approach incorporates a CBF that enforces safety constraints through real-time action filtering.
\item We develop a probabilistic network model to quantify uncertainties arising from sudden hazardous areas, enhancing the anticipation of potential dangers.
\item By employing a proximity-weighted mechanism to integrate neighbors' observations, we utilize collective situational awareness, enabling mutual risk avoidance within the swarm.
\end{enumerate}

We organize the rest of the paper as follows. Section \ref{sec:sample3} formulates the multi-UAV pathfinding problem as a Markov Decision Process. Section \ref{sec:sample4} describes our MARL-based pathfinding approach in detail. Section \ref{sec:sample5} provides the experiments conducted to evaluate our method. Section \ref{sec:sample6} presents our conclusions.

\section{Decentralized Partial Observable Markov Decision Process for Multi-UAV Pathfinding}
\label{sec:sample3}

\subsection{Environment Description}
\label{sec:sample3A}
This study considers the multiple UAV pathfinding problem in a stochastic environment where hazardous areas occur randomly. The objective is to compute collision-free paths for $I$ UAVs in a 3D space, minimizing total path length subject to constraints on avoiding hazardous areas defined by non-convex sets and ensuring inter-UAV distances remain above a safety threshold. The environment state $\boldsymbol{s}$ incorporates the information of all UAVs and hazards.

Let $\mathbb{R}$ and $\mathbb{Z}$ denote the sets of real numbers and integers, respectively. $ \boldsymbol e_i \in \mathbb Z^3, i = 1, 2, 3 $ denotes the standard unit vector. We model the UAV dynamics as follows:
\begin{align}
   \boldsymbol{p}_{u,i}^{[k+1]} =  \boldsymbol{p}_{u,i}^{[k]} + \boldsymbol{v}_{u,i}^{[k]} \tau, \quad \boldsymbol{v}_{u,i}^{[k+1]} =  \boldsymbol{v}_{u,i}^{[k]} + \boldsymbol{a}_{u,i}^{[k]} \tau,
   \label{equ_position}
\end{align}
where $ \boldsymbol{p}_{u,i}^{(t)} \in \mathbb{R}^3 $, $ \boldsymbol{v}_{u,i}^{(t)} \in \mathbb{R}^3 $, and $ \boldsymbol{a}_{u,i}^{(t)} \in \mathbb{R}^3 $ denote the position, velocity, and acceleration of the $i$-th UAV in continuous time $ t $, respectively. These variables are sampled at a time interval $ \tau \in \mathbb{R} $ as $ \boldsymbol{p}_{u,i}^{[k]} := \boldsymbol{p}_{u,i}^{(k\tau)} = \boldsymbol{p}_{u,i}^{(\lfloor t / \tau \rfloor \tau)}$, where $\lfloor t / \tau  \rfloor =\max \left\{ k \in \mathbb{Z}|k \leqslant t/\tau \right\}$.

The location and number of hazardous areas change randomly at regular intervals, and these areas are assumed to not overlap with each other or with any UAV. The positions of newly generated hazardous areas are sampled from a Gaussian mixture model \cite{yang2024gaussian}. We define three distinct high-risk zones, each represented by a 3D Gaussian distribution. When a new hazard is generated, one of these three distributions is chosen uniformly at random, and the hazard's center is then sampled from that chosen distribution.
To address hazardous areas with arbitrary shapes and time-varying geometries, we model the $j$-th hazard as a collection of overlapping spheres, approximating the hazard's point set $\mathbb{H}_j^{[k]}  \approx \bigcup_{m=1}^{M_j} \mathcal{S}(\boldsymbol{p}_{j,m}^{[k]}, \rho_{j,m}^{[k]}) \subseteq \mathbb{R}^3$.
Here, $\mathcal{S}(\boldsymbol{p}_{j,m}^{[k]}, \rho_{j,m}^{[k]}) = \{\boldsymbol{q} \in \mathbb{R}^3 \mid \|\boldsymbol{q} -\boldsymbol{p}_{j,m}^{[k]}\| \leqslant \rho_{j,m}^{[k]}\}$ represents the $m$-th sphere, where $\boldsymbol{p}_{j,m}^{[k]}$ and $\rho_{j,m}^{[k]}$ denote its center and radius, and $M_j$ is the number of spheres. $\boldsymbol{q}$ denotes any point in 3D space being tested to see if it falls inside the sphere. The spheres are generated by clustering $\mathbb{H}_j^{[k]}$ into $M_j$ sub-regions using K-means \cite{liu2023transforming}, with center $\boldsymbol{p}_{j,m}^{[k]} = \sum_{\boldsymbol{q} \in \mathcal{C}_{j,m}^{[k]}} \boldsymbol{q} / M_j$ and radius $\rho_{j,m}^{[k]} = \max_{\boldsymbol{q} \in \mathcal{C}_{j,m}^{[k]}} \|\boldsymbol{q} - \boldsymbol{p}_{j,m}^{[k]}\|$. $\mathcal{C}_{j,m}^{[k]}$ is the point set of the $m$-th cluster. Each sphere's radius evolves with temporal correlation:
\begin{equation}
\rho_{j,m}^{[k+1]} = \rho_{j,m}^{[k]} + \Delta \rho_{j,m}^{[k]} , \quad \Delta \rho_{j,m}^{[k]} \sim \mathcal{N}(\mu_{j,m}, \sigma_{j,m}^2),
\label{eq:radius_dynamics}
\end{equation}
where $\mathcal{N}$ denotes the Gaussian distribution, with the mean $\mu_{j,m}$ and variance $\sigma^2_w$ modeling expansion. 

The observation vector of each UAV is described as
\begin{equation}
\boldsymbol{z}^{[k]}_i = \left( \boldsymbol{p}^{[k]}_{u,i}, \boldsymbol{v}^{[k]}_{u,i}, \{\boldsymbol{p}^{[k]}_{u,j}, \boldsymbol{v}^{[k]}_{u,j}\}_{j \in \mathcal{G}_i}, \{\boldsymbol{p}^{[k]}_{j,m}, \boldsymbol{v}^{[k]}_{j,m}\}_{m \in \mathcal{G}_i} \right),
\label{equ16}
\end{equation}
where $\mathcal{G}_i$ represents the gathering of $N$ nearest neighbors to the $i$-th UAV, including both other UAVs and hazard spheres.

\subsection{CBF-Constrained Action Filtering}
\label{sec:sample3B}
The action vector of each UAV is defined as
\begin{equation}
\boldsymbol{x}_i^{[k]} = (\omega_{\psi,i}^{[k]}, \omega_{\theta,i}^{[k]}, \omega_{\varphi,i}^{[k]}).
\label{equ161}
\end{equation}
It represents the angular velocities for heading, pitch, and roll, and is utilized to calculate the thrust acceleration as follows:
\begin{align}
\boldsymbol{a}^{[k]}_{r,i} = &v_{\text{nom}} \begin{pmatrix}
-s_{\psi} c_{\theta} \omega_{\psi} - c_{\psi} s_{\theta} \omega_{\theta} \\
c_{\psi} c_{\theta} \omega_{\psi} - s_{\psi} s_{\theta} \omega_{\theta} \\
c_{\theta} \omega_{\theta}
\end{pmatrix}  + \begin{pmatrix}
c_{\varphi} c_{\theta} \omega_{\varphi} - s_{\varphi} s_{\theta} \omega_{\theta} \\
s_{\varphi} c_{\theta} \omega_{\varphi} + c_{\varphi} s_{\theta} \omega_{\theta} \\
0
\end{pmatrix}, \label{eq:attitudecalculate}
\end{align}
where $s_{\psi} = \sin(\psi_i^{[k]})$, $c_{\psi} = \cos(\psi_i^{[k]})$, $s_{\theta} = \sin(\theta_i^{[k]})$, $c_{\theta} = \cos(\theta_i^{[k]})$, $s_{\varphi} = \sin(\varphi_i^{[k]})$, $c_{\varphi} = \cos(\varphi_i^{[k]})$, $\omega_{\psi} = \omega_{\psi,i}^{[k]}$, $\omega_{\theta} = \omega_{\theta,i}^{[k]}$, $\omega_{\varphi} = \omega_{\varphi,i}^{[k]}$. $v_{\text{nom}}$ is the nominal speed of the UAV. $\psi_i^{[k]}, \theta_i^{[k]}, \varphi_i^{[k]}$ denote the heading, pitch, and roll angles, respectively. The acceleration $\boldsymbol{a}^{[k]}_{u,i} = \boldsymbol{a}^{[k]}_{r,i}$ when gravity and aerodynamic effects are neglected. Subsequently, using (\ref{equ_position}), the $i$-th UAV transitions to the next position $\boldsymbol{p}_{u,i}^{[k+1]}$, completing the state update.

To provide collision avoidance among UAVs we introduce CBF to provide formal safety guarantees. For the $i$-th UAV and its neighbor $n \in \mathcal{G}_i$, we define the following function:
\begin{equation}
H_{in}^{(t)}\left(\boldsymbol{p}_{u,i}^{(t)}, \boldsymbol{p}_n^{(t)} \right) = \|\boldsymbol{p}_{u,i}^{(t)} - \boldsymbol{p}_n^{(t)}\| - (d_{in}^{(t)} + d_s).
\label{equ24}
\end{equation}
Here, the symbol $\lVert \cdot \rVert$ represents the Euclidean norm. The safe distance $d_s$ establishes a buffer zone around hazards and other UAVs. 
The collision threshold $d_{in}^{(t)}$ depends on the type of neighbor $n$: if the neighbor is another UAV (indexed by $j$), then $d_{ij}^{(t)} :=  \rho_{u,i} + \rho_{u,j}$; if the neighbor is a hazard sphere (indexed by $m$), then $d_{im}^{(t)}:= \rho_{u,i} + \rho_{j,m}^{(t)}$.
The condition $H_{in}^{(t)}\left(\boldsymbol{p}_{u,i}^{(t)}, \boldsymbol{p}_n^{(t)} \right) \geqslant 0$ defines the safe set $\mathcal{C}_{in}$ where the distance constraint is satisfied.

To guarantee that the system state remains within the safe set $\mathcal{C}_{in}$, the standard continuous-time CBF condition requires that the control input $\boldsymbol{x}_i^{(t)}$ satisfies
\begin{align}
\dot{H}_{in}^{(t)} \left(\boldsymbol{p}_{u,i}^{(t)}, \boldsymbol{p}_n^{(t)} \right) \geqslant -\alpha H_{in}^{(t)} \left(\boldsymbol{p}_{u,i}^{(t)}, \boldsymbol{p}_n^{(t)} \right),
\label{equ25}
\end{align}
where $\alpha > 0$ is a design parameter, and $\dot{H}_{in}^{(t)}$ is the time derivative of $H_{in}^{(t)}$ along the system dynamics. This constraint sustains that the system remains within the safe set \cite{ames2019control}. Based on the definition in (\ref{equ24}), this derivative is 
\begin{equation} \dot{H}_{in}^{(t)} \left(\boldsymbol{p}_{u,i}^{(t)}, \boldsymbol{p}_n^{(t)} \right) \approx \frac{(\boldsymbol{p}_{u,i}^{(t)} - \boldsymbol{p}_n^{(t)})^\top (\boldsymbol{v}_{u,i}^{(t)} - \boldsymbol{v}_n^{(t)})}{\|\boldsymbol{p}_{u,i}^{(t)} - \boldsymbol{p}_n^{(t)}\|}. \label{eq:h_dot_def} \end{equation}
The derivation of (\ref{eq:h_dot_def}) is detailed in Appendix-A.
Since our system operates in discrete time with sampling interval $\tau$, we enforce the CBF constraint at each discrete time step $k$. Specifically, we require the chosen action $\boldsymbol{x}_i^{[k]}$ to satisfy the instantaneous derivative condition evaluated at time $k$:
\begin{align}
 \dot{H}_{in}^{[k]} \left(\boldsymbol{p}_{u,i}^{[k]}, \boldsymbol{p}_n^{[k]}\right)
\geqslant -\alpha H_{in}^{[k]} \left(\boldsymbol{p}_{u,i}^{[k]}, \boldsymbol{p}_n^{[k]} \right).
\label{equ25_0}
\end{align}
Here, $\dot{H}_{in}^{[k]}$ is computed using (\ref{eq:h_dot_def}) with the positions and velocities at time $k$. This condition ensures that the rate of change of the barrier function at the current instant points towards the safe region. In practice, we choose the decay rate $\alpha$ within the interval $\left[\alpha_{\min}, \alpha_{\max}\right]$ derived in Appendix-B, which guarantees both the instantaneous CBF inequality and discrete-time forward invariance.

During execution, a safety filter projects the MARL policy's action $\boldsymbol{x}_i^{[k]}$ onto the set of actions satisfying the constraint (\ref{equ25}) for all neighbors $n \in \mathcal{G}_i$. This is formulated as a quadratic program (QP):
\begin{align}
\boldsymbol{x}_{i,s}^{[k]} = \arg&\min_{\boldsymbol{\hat{x}}_i^{[k]}} \|\boldsymbol{\hat{x}}_i^{[k]} - \boldsymbol{x}_i^{[k]}\|^2, \nonumber \\ \text{s.t.} \quad \dot{H}_{in}^{[k]} \left(\boldsymbol{p}_{u,i}^{[k]}, \boldsymbol{p}_n^{[k]}\right) &\geqslant -\alpha H_{in}^{[k]} \left(\boldsymbol{p}_{u,i}^{[k]}, \boldsymbol{p}_n^{[k]} \right), \forall n \in \mathcal{G}_i.
\label{equ26}
\end{align}
This QP finds the safe action $\boldsymbol{x}_{i,s}^{[k]}$ that is closest to the original proposed action $\boldsymbol{x}_{i}^{[k]}$. During execution, the original action is checked first, and the QP is solved only if necessary. If the QP is infeasible, a fallback action $\boldsymbol{x}_{i,s}^{[k]} = 0$ is used, causing the UAV to hover. This ensures UAVs remain within safe operational bounds.

\subsection{Reward Designing}
\label{sec:sample3C}
We design a multi-component reward function that balances goal-reaching against risk-aversion. This reward is composed of the following three components:
\begin{equation}
r^{[k]} = r_{g}^{[k]} + \epsilon_t r_{t}^{[k]} + r_{c}.
\label{equ20}
\end{equation}
The goal-reaching reward $r_g^{[k]}$ is defined as
\begin{equation}
r_{g}^{[k]} = - \sum_{i=1}^I\frac{\| \boldsymbol{p}_{i,g} - \boldsymbol{p}_{u,i}^{[k+1]} \|}{\| \boldsymbol{p}_{i,g} - \boldsymbol{p}_{u,i}^{[0]} \|},
\label{eq:reward_goal}
\end{equation}
where $\boldsymbol{p}_{i,g}$ is the target position of the $i$-th UAV.
To encourage safe navigation, we define a threat zone around each neighbor $n \in \mathcal{G}_i$ with a distance $d_t > d_{in}^{[k]}$. A linearly scaled penalty $r_{in}^{[k]}$ is applied only when the UAV enters this zone:
\begin{equation}
\begin{array}{c}
 r_{in}^{[k]} =
\left\{ \begin{array}{c}
\displaystyle -\frac{d_{t} - p_{in}^{[k+1]}}{d_{t} - d_{in}^{[k+1]}}, \  \text{if } d_{in}^{[k+1]} < p_{in}^{[k+1]} < d_t \\
0, \  \text{otherwise} \\ \end{array} \right. ,
\end{array}
\label{eq:reward_safe}
\end{equation}
where $p_{in}^{[k+1]}=\| \boldsymbol{p}_{u,i}^{[k+1]} - \boldsymbol{p}_{n}^{[k+1]} \|$. The threat penalty $r_{t}^{[k]}$ is calculated as
\begin{equation}
\begin{array}{c}
r_{t}^{[k]} = \displaystyle \sum_{i=1}^I\sum_{n \in \mathcal{G}_i} r_{in}^{[k]}.
\end{array}
\label{eq:reward_safe}
\end{equation}
It discourages entry into the warning zone without overpowering the goal-seeking objective in safe regions. The hyperparameter $\epsilon_t$ allows for explicit tuning of the agent's risk-aversion.
Finally, a constant penalty $r_{c} \neq 0$ if the collision happens, otherwise, $r_{c}= 0$. 

\section{MARL-Based Pathfinding Approach}
\label{sec:sample4}
\begin{figure*}[t]
    \centering
    \includegraphics[width=0.99\textwidth]{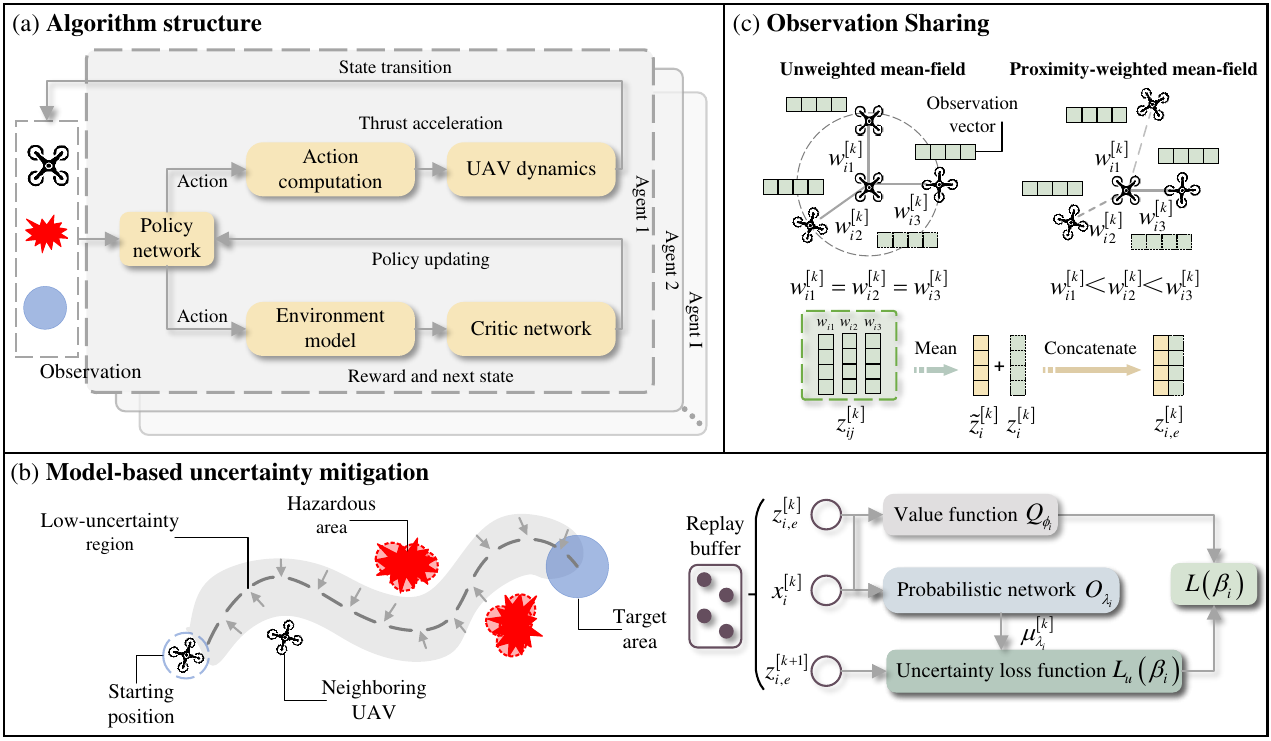}
    \caption{Overview of the proposed method. (a) Framework of RL-based decision-making for UAV pathfinding. (b) Illustration of model-based uncertainty mitigation during policy learning. (c) Comparison of unweighted mean-field and proximity-weighted mean-field in extended observations.}
   \label{fig1}
\end{figure*}

In this section, we present our training and execution scheme in MARL. The framework of our approach is illustrated in Fig. \ref{fig1}(a).

\subsection{Uncertainty Mitigation in Policy Training}
\label{sec:sample4A}
In dynamic environments with sudden hazardous areas, UAVs must not only respond to current observations but also anticipate future uncertainties to achieve rapid and safe pathfinding. Traditional RL methods often neglect these uncertainties, resulting in fragile policies that falter under abrupt changes. To address this, we propose a model-based approach that quantifies uncertainty using a PNN \cite{huang2024practical}. Unlike a standard network that produces a deterministic output, the PNN outputs its result as a probability distribution-specifically, a Gaussian distribution $ \mathcal{N}(\mu, \varSigma) $ in this study. The term $\mu$ is the mean of this predicted Gaussian distribution, representing the most likely or expected output. The term $\varSigma$ is the covariance of that same distribution, which quantifies the uncertainty around that expected output.

The core idea involves adopting the PNN as virtual environment models to approximate the observation transitions of UAVs, enabling each agent to predict future observations while accounting for uncertainty.
We model the observation transition for each UAV using a PNN $ O_{\lambda_i} $, which takes the current observation $ \boldsymbol{z}_i^{[k]} $ and action $ \boldsymbol{x}_{i,s}^{[k]} $ as inputs to predict the next observation $ \boldsymbol{\hat{z}}_i^{[k+1]} = O_{\lambda_i} \left( \boldsymbol{z}_i^{[k]}, \boldsymbol{x}_{i,s}^{[k]} \right) $. Here, $ \lambda = (\lambda_1, \dots, \lambda_I) $ denotes the collection of network parameters across all $ I $ agents. The observation $\boldsymbol{z}_i^{[k]}$ for the $i$-th UAV represents a vector of real-world states, including its own position and velocity, as well as those of nearby UAVs and hazardous spheres, as detailed in (\ref{equ16}). The safety-filtered action $\boldsymbol{x}_{i,s}^{[k]}$, obtained after applying the CBF constraint as described in (\ref{equ26}), corresponds to the angular velocities that determine the UAV's actual thrust acceleration and subsequent motion. The network parameterizes a Gaussian distribution over the next observation: $ f\left( \boldsymbol{z}_i^{[k+1]} \mid \boldsymbol{z}_i^{[k]}, \boldsymbol{x}_{i,s}^{[k]} \right) = \mathcal{N} \left( \mu_{\lambda_i}^{[k]}, \varSigma_{\lambda_i}^{[k]} \right) $. The mean $ \mu_{\lambda_i}^{[k]} $ represents the expected next observation for path guidance, while the covariance $ \varSigma_{\lambda_i}^{[k]} $ quantifies uncertainty arising from hazards. This concept is visualized in Fig. \ref{fig1}(b), where dashed curves represent the mean and shaded regions depict the covariance. A Softplus activation ensures the predicted covariance $ \varSigma_{\lambda_i}^{[k]} $ remains strictly positive.

These networks are trained by maximizing the likelihood of transitions sampled from the replay buffer, using the negative log-likelihood loss: 
\begin{align} L\left(\lambda_i\right) = E \left( \frac{\log \varSigma_{\lambda_i}^{[k]} }{2} + \frac{\left( \boldsymbol{z}_i^{[k+1]} - \mu_{\lambda_i}^{[k]} \right)^2}{2 \varSigma_{\lambda_i}^{[k]} } \right), \label{equ46} \end{align} 
where $E$ denotes the mathematical expectation. This loss function promotes accurate distribution fitting, encouraging low covariance in stable regions and higher covariance in areas where hazards introduce unpredictability.

Then, we introduce a constrained uncertainty loss that penalizes deviations from the predicted mean, guiding UAVs toward regions of low uncertainty:
\begin{align}
\displaystyle L_{u}\left(\beta_i\right) = E \lVert \boldsymbol{z}_i^{[k+1]} - \mu_{\lambda_i}^{[k]} \lVert.
\label{equ47}
\end{align}
It encourages the $i$-th UAV to minimize deviations from the predicted mean, steering it toward regions with predictable dynamics. Since areas of high uncertainty, such as abrupt hazards, inherently lead to larger prediction errors, this loss implicitly guides the policy to avoid those high-uncertainty regions. We utilize the multi-agent deep deterministic policy gradient (MADDPG) algorithm \cite{chai2024multi} to train the UAV's policy.
It provides an individual policy network $\pi_{\beta_i}\left( \boldsymbol{z}^{[k]}_i \right)$ for the $i$-th UAV, allowing the agent to determine an action based on its local observation during execution. The subscript $\beta=\left(\beta_1, \dots, \beta_I\right)$ represents the weighting parameter of policy networks.
We refine the loss function to balance high rewards and low uncertainty, described as
\begin{align}
\displaystyle L\left(\beta_i\right) = L^*\left(\beta_i\right) + \epsilon_u L_{u}\left(\beta_i\right),
\label{equ48}
\end{align}
where $\epsilon_u$ balances reward maximization and uncertainty minimization. $L^*\left(\beta_i\right)$ is the standard loss function for the policy training in MADDPG, given by
\begin{equation}
    \begin{array}{c}
        L^*\left(\beta_i\right)=E\left(-Q_{\phi_i}\left( \boldsymbol{z}^{[k]},\boldsymbol{x}_{s}^{[k]}) \right)\right),
    \end{array}\label{equ4}
\end{equation}
where $Q_{\phi_i}(\boldsymbol{z}^{[k]},\boldsymbol{x}_{s}^{[k]} )$ is an
observation-action value function, producing gradients to optimize the policy. $\phi=\left(\phi_1, \dots, \phi_I\right)$ represents the weighting parameter of value-function networks.
This centralized function is iteratively fitted using the following loss function:
\begin{equation}
    \begin{array}{c}
        L\left( \phi_{i} \right)=E\lVert Q_{\phi_i}\left( \boldsymbol{z}^{[k]},\boldsymbol{x}_{s}^{[k]} \right) -y_{i}^{[k]} \rVert ^2.
    \end{array}\label{equ1}
\end{equation}
The $Q$-target $y_{i}^{[k]}=r^{[k]} +\gamma \max Q_{\phi_i}\left( \boldsymbol{z}^{[k+1]},\boldsymbol{x}_{s}^{[k+1]} \right)$, where $\gamma$ denotes the discount factor.

\begin{algorithm*}[t]
\renewcommand{\algorithmicrequire}{\textbf{Input:}}
\renewcommand{\algorithmicensure}{\textbf{Initialization:}}
\caption{MARL-Based Pathfinding Approach}
\label{alg2}
\begin{algorithmic}[1]
\REQUIRE Training episodes ($E_k$); Steps per episode ($K$); Number of agents ($I$); \\
\ENSURE Replay buffer; Network parameters: critics $\phi$, policies $\beta$, uncertainty models $\lambda$;
\FOR {$e = 1$ \TO $E_k$} 
    \STATE Initialize environment and get initial joint observation $\boldsymbol{z}^{[0]}$;
    \FOR {$k = 0$ \TO $K$} 
        \FOR {$i = 1$ \TO $I$} 
            \STATE Integrate neighbors' observations $\boldsymbol{\bar{z}}_{i}^{[k]} \gets$ Eq.~(\ref{equ32}); \COMMENT{Proximity-weighted mean-field observation sharing}
            \STATE Generate action $\boldsymbol{x}_i^{[k]} \gets \pi_{\beta_i} \left( \boldsymbol{z}_{i}^{[k]}, \boldsymbol{\bar{z}}_{i}^{[k]} \right)$;
            \STATE Correct action if unsafe: $\boldsymbol{x}_{i,s}^{[k]} \gets$ Eq.~(\ref{equ26}); \COMMENT{Enforce collision-free paths via CBF}
        \ENDFOR
        \STATE Execute safe joint action $\boldsymbol{x}_s^{[k]}$ in environment to get reward $r^{[k]}$ and next joint observation $\boldsymbol{z}^{[k+1]}$;
        \STATE Store transition in replay buffer: $(\boldsymbol{z}^{[k]}, \boldsymbol{x}_{s}^{[k]}, r^{[k]}, \boldsymbol{z}^{[k+1]})$;
        \STATE Sample minibatch of transitions from replay buffer; \COMMENT{Prepare for centralized updates}
        \FOR {$i = 1$ \TO $I$} 
            \STATE Update critic: $L(\phi_{i}) \gets$ Eq.~(\ref{equ1}); 
            \STATE Update uncertainty model: $L(\lambda_i) \gets$ Eq.~(\ref{equ46}); \COMMENT{Update PNN for uncertainty quantification}
            \STATE Compute uncertainty loss: $L_u(\beta_i) \gets$ Eq.~(\ref{equ47}); \COMMENT{Mitigate uncertainty in policy training}
            \STATE Update policy: $L(\beta_i) \gets$ Eq.~(\ref{equ48}); \COMMENT{Balance rewards and uncertainty}
        \ENDFOR
    \ENDFOR
\ENDFOR
\end{algorithmic}
\end{algorithm*}

\subsection{Integrating Neighbors' Observations into Planning}
\label{sec:sample4B}

To integrate observations on neighboring UAVs, we employ a proximity-weighted mean-field mechanism that evaluates interaction strengths among UAVs based on proximity, as shown in Fig. \ref{fig1}(c). The mean-field MARL approach \cite{wang20223m} approximate the influence of neighbors using average interaction terms: $Q_i\left( \boldsymbol{z}^{[k]},\boldsymbol{x}^{[k]} \right) = \sum_{j \in \mathcal{G}_i} Q_i\left( \boldsymbol{z}_i^{[k]},\boldsymbol{z}_j^{[k]},\boldsymbol{x}_i^{[k]},\boldsymbol{x}_j^{[k]} \right) \big/ N$. Inspired by this principle of representing collective effects through aggregation, we express the policy for the $i$-th agent considering pairwise interactions with its neighbors in a proximity-weighted form:
\begin{align}
\pi_i\left( \boldsymbol{z}^{[k]} \right) &=  \sum_{j \in \mathcal{G}_i} w_{ij}^{[k]} \pi_i\left( \boldsymbol{z}_i^{[k]}, \boldsymbol{z}_j^{[k]} \right) \nonumber \\
&\approx \pi_i\left( \boldsymbol{z}_i^{[k]},\boldsymbol{\bar{z}}_i^{[k]} \right),
\label{equ38}
\end{align}
where
\begin{align}
        \displaystyle \boldsymbol{\bar{z}}_i^{[k]}=\sum_{j\in \mathcal{G}_i}{w_{ij}^{[k]}\boldsymbol{z}_j^{[k]}}, \quad w_{ij}^{[k]}=\frac{ 1 \big/ \lVert \boldsymbol{p}_{u,i}^{[k]}-\boldsymbol{p}_{u,j}^{[k]} \rVert}{\sum_{l\in \mathcal{G}_i} \left(1 \big/ \lVert \boldsymbol{p}_{u,i}^{[k]}-\boldsymbol{p}_{u,l}^{[k]} \rVert\right)}.
        \label{equ32}
\end{align}
The symbol $w_{ij}^{[k]}$ represents the interaction weight between the $i$-th UAV and the $j$-th UAV, based on their distance, with $\sum_{j\in \mathcal{G}_i} w_{ij}^{[k]}=1$. Closer proximity leads to a stronger interaction weight, highlighting a greater potential for collision, and thus increasing the focus on that neighbor. The proof of the approximation in (\ref{equ38}) is provided in Appendix-C.

Compared with a unweighted mean-field approach \cite{wang20223m} that weights all neighbors equally, our method is more physically intuitive for swarm robotics. It explicitly prioritizes information from closer UAVs, which is critical for immediate collision avoidance and localized coordination where nearby agents have the most significant impact. While a more complex attention mechanism \cite{hao2023gat} could learn similar weightings, our method provides this crucial inductive bias directly from physical principles, avoiding the additional model complexity, potential training instability, and computational overhead of a learned attention module. This design provides a middle ground, offering a more informed aggregation than unweighted mean-field without the costs of a fully learned mechanism.

Acquiring these neighbors' observations relies on inter-agent communication, where each UAV broadcasts its local observation vector to its neighbors. While this assumes a communication channel, the overhead is kept manageable as the transmitted messages contain low-dimensional state information rather than high-bandwidth raw sensor data like images or point clouds. The robustness of our approach under non-ideal communication conditions, such as message loss and sensor noise, are then experimentally validated in Section \ref{sec:sample5E}. The framework of the MARL-based pathfinding approach is illustrated in Algorithm \ref{alg2}.

\section{EXPERIMENTS}
\label{sec:sample5}

In this section, we present extensive experiments to evaluate the performance of our method for pathfinding in stochastic environments with abrupt hazardous areas. 

\subsection{Setting Up}
\label{sec:sample5A}

\begin{figure}[t]
    \centering    \includegraphics[width=0.48\textwidth]{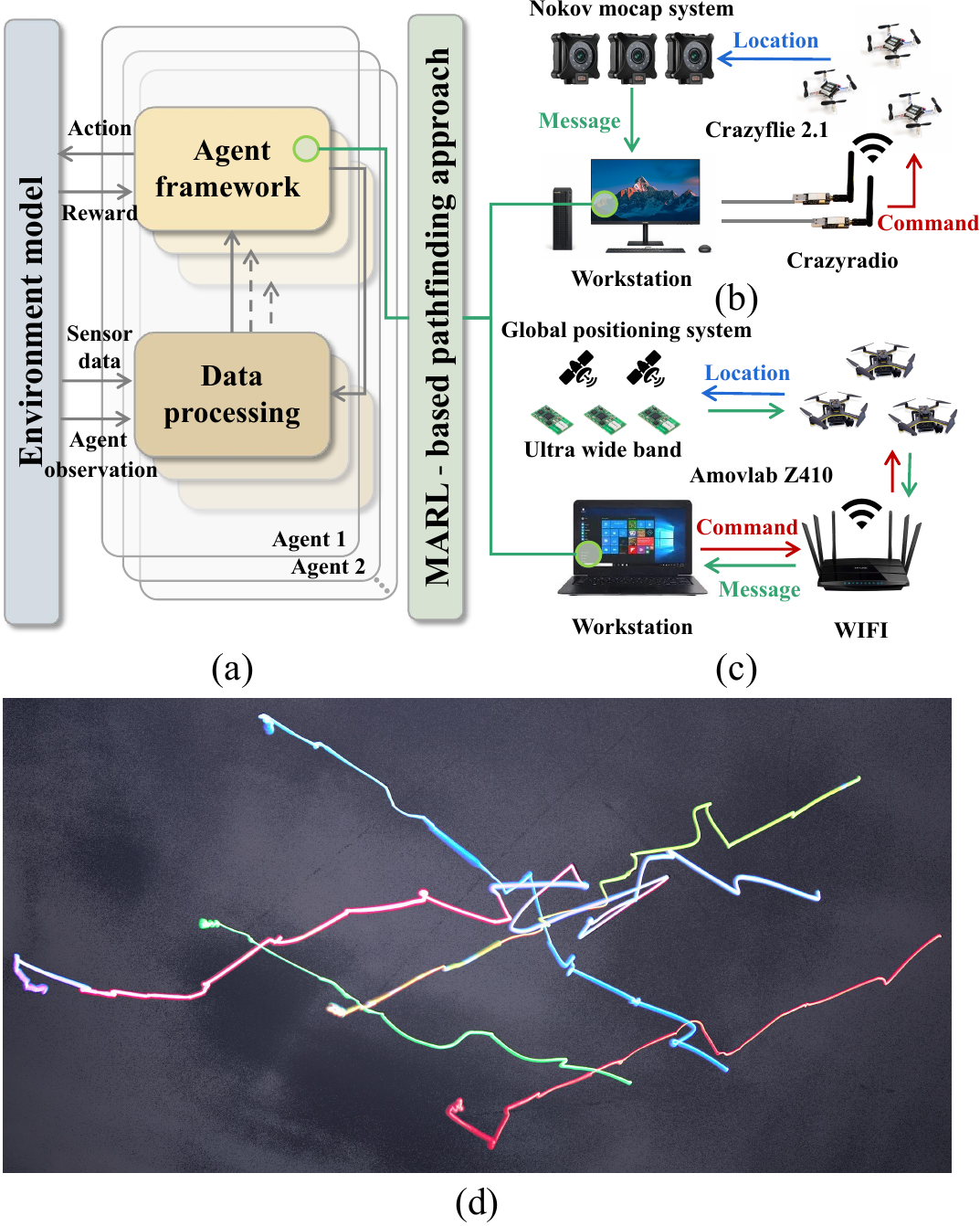}
    \caption{Experiment and implementation setup. (a) Simulations. (b) Indoor experiments. (c) Outdoor experiments. (d) Long-exposure photo of a flight with multiple UAVs.}
    \label{fig4_0}
\end{figure}

\begin{table}[t]
\renewcommand{\arraystretch}{1.08}
\setlength{\tabcolsep}{10pt}
\centering
\caption{Parameter Settings.}
\begin{tabular}{c c}
\toprule
Parameters & Values \\
\midrule
Episode number $E_k$ & 60 \\
Episode length $K$ & 500 \\
Time interval $\tau$ & 0.1 s \\
UAV radius $\rho_{u,i}$ & 0.5 m\\
Hazard sphere radius $\rho_{j,m}$ & [0.5, 1.5] m\\
Hazards number, Sphere number $M_j$ & 10, 5\\
Mean $\mu_{j,m}$, Variance $\sigma^2_w$ & 0.1 m/s, 0.05 m \\
Nominal speed $v_\text{nom}$ & 3 m/s \\
Observed neighbor number $N$ & 3\\
Collision penalty $r_{c}$ & -500\\
Threat distance for reward $d_{t}$ & 4 m\\
Safe distance for CBF $d_s$ & 1 m\\
Discount factor $\gamma$ & 0.99\\
Batch size & 1256\\
Memory size & 1e5\\
Actor/Critic learning rate & 1e-3\\
Soft update weight & 0.01\\
Action limits & [-1.57, 1.57] rad/s\\
Weight $\epsilon_t$ & 1.0\\
Weights $\epsilon_u$ & [0.2, 0.8]\\
\bottomrule
\end{tabular}
\label{table1}
\end{table}

\begin{table}[t]
\renewcommand{\arraystretch}{1.08} 
\setlength{\tabcolsep}{5pt}
\centering
\caption{Network Architectures}
\begin{tabular}{c c c c c}
\toprule
Network & Input Shape & Hidden Layers & Output Shape \\
\midrule
Policy  & $m_o$         & 4 layers, 128 units each & $m_a$        \\
Critic  & $m_{o}+m_{a}$     & 4 layers, 256 units each & 1       \\
PNN  & $m_o+m_a$     & 4 layers, 256 units each & 2$m_o$        \\
\bottomrule
\end{tabular}
\label{table:net_arch}
\flushleft
\footnotesize{\textit{Note:}  All networks are multi-layer perceptrons and all hidden layers use the ReLU activation function. The policy network's output layer uses a Tanh activation function. The PNN's output parameterizes a Gaussian distribution over the next state.}
\end{table}

\begin{figure*}[t]
    \centering    \includegraphics[width=0.99\textwidth]{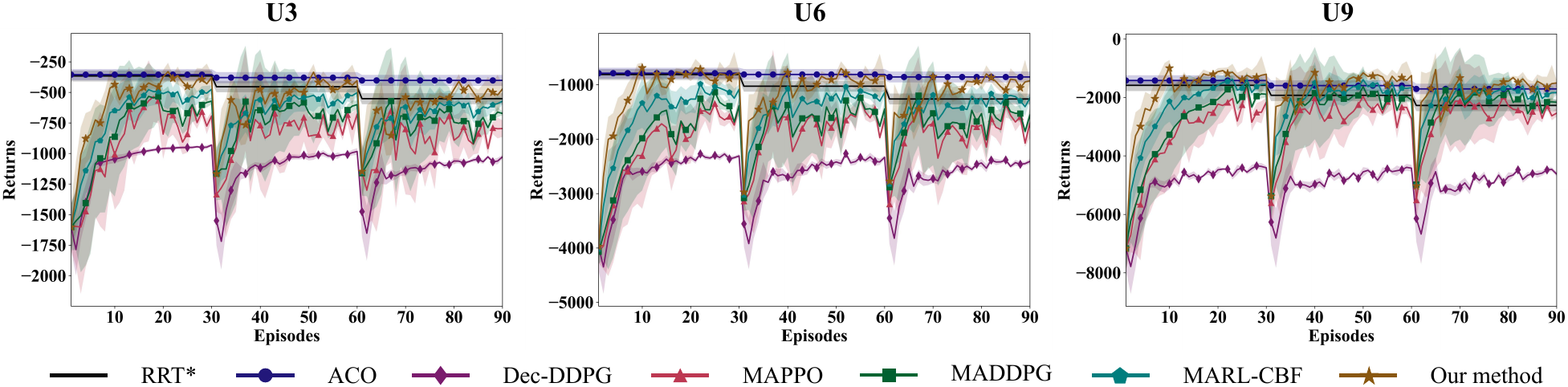}
    \caption{Learning curves of baselines and our Method. The figures plot episode returns versus the number of training episodes. Solid curves represent the mean values across all instances, and the shaded regions indicate the standard deviation.}
    \label{fig2}
\end{figure*}

Our multi-UAV simulation platform, developed in Python, provides a high-fidelity environment by incorporating real-world physical factors such as gravity and aerodynamic effects, detailed in Appendix-D. As depicted in Fig. \ref{fig4_0}(a), the simulation operates on the agent-environment feedback loop central to our MARL approach. In each time step, a UAV receives raw sensor data from the environment model, which is processed to derive the agent's current observation. Based on this observation, the agent's policy selects an action to execute, prompting the environment to return a reward signal and transition to a new state, thus completing the loop. The accompanying web-based visualizations are implemented in Java. All experiments are conducted within this platform in a 100 m$\times$100 m$\times$20 m 3D space, featuring scenarios with 3, 6, and 9 UAVs (U3, U6, and U9) and dynamic hazardous areas changing at frequencies of 0.05, 0.1, and 0.2 Hz (F0.05, F0.1, F0.2). We define three distinct high-risk zones for the hazard generation, with centers at $\mu_1 =$ (25, 75, 10) m, $\mu_2 =$ (75, 50, 10) m, $\mu_3 =$ (25, 25, 10) m. Each zone's distribution used a diagonal covariance of $\varSigma =$ diag(15$^2$, 15$^2$, 5$^2$) m$^2$. Additional parameters are detailed in Table \ref{table1}, and the specific network architectures are provided in Table \ref{table:net_arch}. The rationale for the selection of key parameters is detailed in Appendix-F.

\begin{table*}[t]
\renewcommand{\arraystretch}{1.08}
\setlength{\tabcolsep}{3pt}
\centering
\caption{Performance Comparison of Algorithms Across Varying Swarm Sizes and Hazard Frequencies.}
\label{table2}
\begin{threeparttable}
\begin{tabular}{c c ccc ccc ccc}
\toprule
\multirow{2}{*}{\makecell{Hazard\\Frequency}} & \multirow{2}{*}{Method} & \multicolumn{3}{c}{U3} & \multicolumn{3}{c}{U6} & \multicolumn{3}{c}{U9}\\
\cmidrule(lr){3-5} \cmidrule(lr){6-8} \cmidrule(lr){9-11}
& & Ti.($\times 10^{-3}$s) & Re. & Ra.($\%$) & Ti.($\times 10^{-3}$s) & Re. & Ra.($\%$) & Ti.($\times 10^{-3}$s) & Re. & Ra.($\%$)\\
\midrule
\multirow{7}{*}{0.05 Hz} & RRT* & 3.37 $\pm$ 0.31 & -365 $\pm$ 43 & 100 $\pm$ 0 & 6.57 $\pm$ 0.62 & -812 $\pm$ 87 & 95 $\pm$ 2 & 13.40 $\pm$ 1.23 & -1583 $\pm$ 152 & 85 $\pm$ 4 \\
& ACO & 5.23 $\pm$ 0.48 & -355 $\pm$ 39 & 100 $\pm$ 0 & 12.93 $\pm$ 1.15 & -788 $\pm$ 82 & 96 $\pm$ 2 & 31.30 $\pm$ 2.54 & -1418 $\pm$ 146 & 87 $\pm$ 3 \\
& Dec-DDPG & \multirow{3}{*}{\textbf{1.63 $\pm$ 0.18}} & -930 $\pm$ 97 & 74 $\pm$ 6 & \multirow{3}{*}{\textbf{3.13 $\pm$ 0.31}} & -2241 $\pm$ 213 & 67 $\pm$ 8 & \multirow{3}{*}{\textbf{6.93 $\pm$ 0.55}} & -4383 $\pm$ 418 & 35 $\pm$ 9 \\
& MAPPO & & -728 $\pm$ 76 & 85 $\pm$ 4 & & -1275 $\pm$ 134 & 83 $\pm$ 5 & & -2331 $\pm$ 224 & 74 $\pm$ 6 \\
& MADDPG & & -607 $\pm$ 64 & 90 $\pm$ 3 & & -1272 $\pm$ 127 & 83 $\pm$ 4 & & -2059 $\pm$ 203 & 78 $\pm$ 5 \\
& MARL-CBF & 2.15 $\pm$ 0.24 & -410 $\pm$ 45 & 99 $\pm$ 1 & 4.08 $\pm$ 0.41 & -855 $\pm$ 92 & 97 $\pm$ 2 & 8.54 $\pm$ 0.78 & -1280 $\pm$ 135 & 92 $\pm$ 3 \\
& Our method & 1.75 $\pm$ 0.19 & \textbf{-359 $\pm$ 36$^{*}$}& \textbf{100 $\pm$ 0$^{*}$}& 3.35 $\pm$ 0.32 & \textbf{-695 $\pm$ 71$^{*}$}& \textbf{99 $\pm$ 1$^{*}$}& 7.28 $\pm$ 0.59 & \textbf{-1091 $\pm$ 108$^{*}$}& \textbf{95 $\pm$ 2$^{*}$}\\
\hline
\multirow{7}{*}{0.1 Hz} & RRT* & 4.83 $\pm$ 0.42 & -453 $\pm$ 51 & 98 $\pm$ 1 & 8.40 $\pm$ 0.73 & -1027 $\pm$ 112 & 91 $\pm$ 3 & 17.40 $\pm$ 1.51 & -1925 $\pm$ 188 & 79 $\pm$ 5 \\
& ACO & 9.70 $\pm$ 0.81 & -379 $\pm$ 44 & 100 $\pm$ 0 & 18.70 $\pm$ 1.62 & -811 $\pm$ 89 & 95 $\pm$ 2 & 49.20 $\pm$ 4.03 & -1592 $\pm$ 164 & 87 $\pm$ 4 \\
& Dec-DDPG & \multirow{3}{*}{\textbf{1.80 $\pm$ 0.19}} & -978 $\pm$ 108 & 73 $\pm$ 7 & \multirow{3}{*}{\textbf{3.47 $\pm$ 0.33}} & -2292 $\pm$ 234 & 65 $\pm$ 8 & \multirow{3}{*}{\textbf{7.63 $\pm$ 0.61}} & -4422 $\pm$ 453 & 31 $\pm$ 9 \\
& MAPPO & & -753 $\pm$ 82 & 85 $\pm$ 4 & & -1508 $\pm$ 157 & 78 $\pm$ 6 & & -2509 $\pm$ 262 & 69 $\pm$ 7 \\
& MADDPG & & -636 $\pm$ 69 & 90 $\pm$ 3 & & -1333 $\pm$ 142 & 80 $\pm$ 5 & & -2253 $\pm$ 237 & 73 $\pm$ 6 \\
& MARL-CBF & 2.41 $\pm$ 0.26 & -495 $\pm$ 55 & 98 $\pm$ 2 & 4.53 $\pm$ 0.45 & -980 $\pm$ 105 & 95 $\pm$ 3 & 9.21 $\pm$ 0.85 & -1550 $\pm$ 162 & 90 $\pm$ 4 \\
& Our method & 1.96 $\pm$ 0.20 & \textbf{-389 $\pm$ 41$^{*}$}& \textbf{100 $\pm$ 0$^{*}$}& 3.72 $\pm$ 0.35 & \textbf{-762 $\pm$ 78$^{*}$}& \textbf{97 $\pm$ 2$^{*}$}& 8.15 $\pm$ 0.66 & \textbf{-1209 $\pm$ 123$^{*}$}& \textbf{93 $\pm$ 3$^{*}$}\\
\hline
\multirow{7}{*}{0.2 Hz} & RRT* & 8.43 $\pm$ 0.71 & -551 $\pm$ 62 & 95 $\pm$ 2 & 13.30 $\pm$ 1.12 & -1262 $\pm$ 138 & 83 $\pm$ 4 & 26.60 $\pm$ 2.23 & -2273 $\pm$ 224 & 71 $\pm$ 6 \\
& ACO & 19.13 $\pm$ 1.54 & -402 $\pm$ 47 & \textbf{100 $\pm$ 0} & 32.87 $\pm$ 2.81 & -857 $\pm$ 96 & 94 $\pm$ 3 & 76.97 $\pm$ 6.05 & -1706 $\pm$ 183 & 82 $\pm$ 5 \\
& Dec-DDPG & \multirow{3}{*}{\textbf{2.20 $\pm$ 0.21}} & -1036 $\pm$ 118 & 71 $\pm$ 8 & \multirow{3}{*}{\textbf{4.23 $\pm$ 0.42}} & -2353 $\pm$ 254 & 60 $\pm$ 9 & \multirow{3}{*}{\textbf{8.90 $\pm$ 0.74}} & -4534 $\pm$ 485 & 25 $\pm$ 10 \\
& MAPPO & & -832 $\pm$ 93 & 83 $\pm$ 5 & & -1595 $\pm$ 172 & 75 $\pm$ 7 & & -2706 $\pm$ 293 & 63 $\pm$ 8 \\
& MADDPG & & -708 $\pm$ 79 & 88 $\pm$ 4 & & -1446 $\pm$ 158 & 78 $\pm$ 6 & & -2434 $\pm$ 264 & 67 $\pm$ 7 \\
& MARL-CBF & 2.85 $\pm$ 0.30 & -610 $\pm$ 70 & 97 $\pm$ 2 & 5.15 $\pm$ 0.53 & -1150 $\pm$ 125 & 92 $\pm$ 4 & 10.95 $\pm$ 0.98 & -1890 $\pm$ 205 & 86 $\pm$ 5 \\
& Our method & 2.43 $\pm$ 0.23 & \textbf{-450 $\pm$ 52$^{*}$}& \textbf{99 $\pm$ 1$^{*}$}& 4.68 $\pm$ 0.45 & \textbf{-833 $\pm$ 88$^{*}$}& \textbf{95 $\pm$ 2$^{*}$}& 9.82 $\pm$ 0.81 & \textbf{-1355 $\pm$ 142$^{*}$}& \textbf{90 $\pm$ 3$^{*}$}\\
\bottomrule
\end{tabular}
\flushleft
\footnotesize{\textit{Note:} The table presents computation time (Ti.), episode returns (Re.), and success rate (Ra.). Data values are reported as mean $\pm$ standard deviation across 10 independent evaluation batches. The symbol $*$ indicates that the improvement of our method over the best-performing baseline is statistically significant with a $p$-value $<$ 0.05 (paired $t$-test). }
\end{threeparttable}
\end{table*}

The algorithm is trained across 100 unique instances, each initialized with a different random seed to vary the starting positions of agents, locations of hazards, and initial network weights. The training curriculum are designed to progressively increase environmental dynamism: the hazard change frequency is set to 0.05 Hz for episodes 1--30, increased to 0.1 Hz for episodes 31--60, and further increased to 0.2 Hz for episodes 61--90. To balance reward maximization with uncertainty mitigation, the parameter $\epsilon_u$ is linearly annealed from an initial value of 0.8 to 0.2 over the course of training. This strategy prioritizes minimizing uncertainty in the early stages before gradually shifting focus to optimizing path efficiency.
Finally, we validate our approach through offline training in the simulation followed by online decision-making in real-world indoor and outdoor environments.

To quantitatively assess the proposed method, we employ three metrics to evaluate the performance:
\begin{enumerate}
    \item Episode Return: The cumulative reward summed over all agents and time steps in an episode, formulated as $R = \sum_{k=0}^K r^{[k]}$. This metric reflects the comprehensive performance considering both task efficiency and safety.
    \item Computation Time: The average wall-clock duration required for the system to compute actions for all UAVs at each time step, verifying real-time capability.
    \item Success Rate: The proportion of evaluation episodes where all UAVs successfully reach their targets without entering hazardous areas or colliding with other agents.
\end{enumerate}

\subsection{Comparative Analysis}
\label{sec:sample5B}
We compare our training approach with several baselines, including decentralized DDPG (Dec-DDPG) \cite{wu2024model}, multi-agent proximal policy optimization (MAPPO) \cite{yan2022}, MADDPG \cite{chai2024multi}, and MARL-CBF \cite{cao2023safe}. Additionally, we evaluate our trained policies against non-learning approaches such as RRT* \cite{guo2023hpo} and ACO \cite{yan2024efficient}. To ensure fair comparisons, all learning-based algorithms use the same network architecture, and all methods are evaluated under identical experimental conditions.

As shown in Fig. \ref{fig2}, non-learning algorithms achieve high returns via iterative optimization in uncertain environments, while learning-based methods start randomly and improve over time. The MARL-CBF baseline, while stable due to its reactive safety shield, converges to lower returns. This demonstrates that a purely reactive safety mechanism can be overly conservative, hindering the agent from discovering efficient paths, thus leading to lower success rates due to task timeouts.
In contrast, our method converges faster than other MARL baselines to a superior policy, achieving returns comparable to the non-learning algorithms. Statistical analysis (paired $t$-test on the final 10 episodes) confirms that this performance superiority over MARL baselines is statistically significant ($p <$ 0.05), validating the effectiveness of the proposed approach. Although our method experiences a brief performance dip when environmental dynamism increases at episodes 31 and 61, it quickly adapts and recovers, showcasing its superior stability.

The quantitative results, averaged over 100 instances, are summarized in Table \ref{table2}. We report the mean computation time per step (Ti.), the final episode returns (Re.), and the success rate (Ra.). While all methods achieve real-time performance (Ti. $<\tau=$ 0.1 s), there are clear distinctions in their efficiency and effectiveness. Classical planners like ACO and RRT* achieve high returns in simpler scenarios but require significantly more computation time and scale poorly as swarm size and hazard frequency increase.

The computation time of the MARL methods warrants a specific discussion. The end-to-end MARL baselines (Dec-DDPG, MAPPO, MADDPG) are the fastest, relying on a single network pass. Our method is slightly slower due to the occasional activation of the CBF safety filter. The MARL-CBF baseline is slower still, as its less-informed policy requires more frequent interventions from the computationally more expensive CBF solver. Despite this, our method's computation time remains significantly lower than the classical planners. Overall, our approach demonstrates a superior trade-off, outperforming all baselines in achieving a high success rate across all scenarios while maintaining low computation time and high reward efficiency.

\begin{figure*}[t]
    \centering    \includegraphics[width=0.99\textwidth]{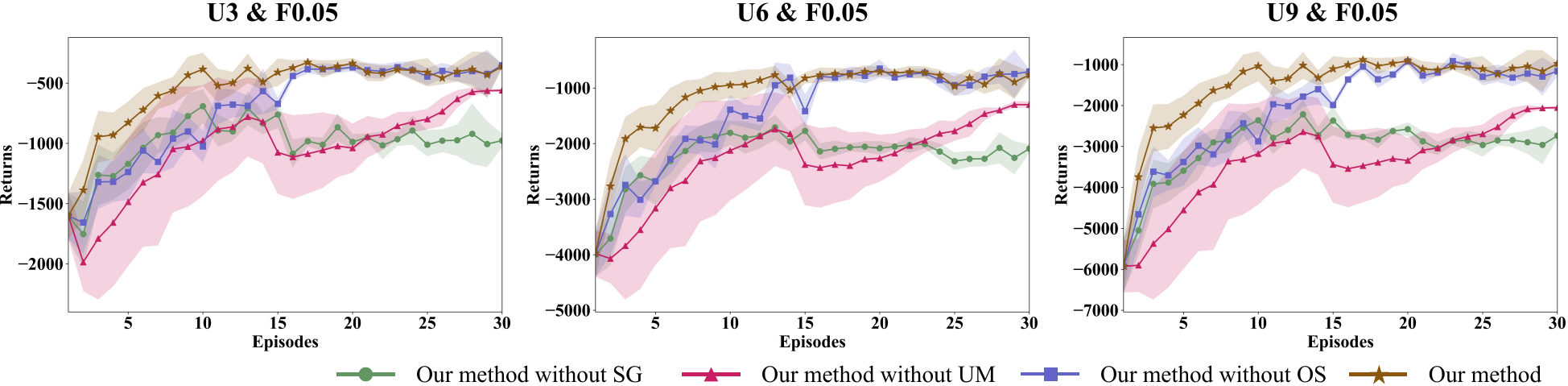}
    \caption{Ablation study results of our method.}
    \label{fig3}
\end{figure*}

\begin{figure*}[t]
    \centering    \includegraphics[width=0.99\textwidth]{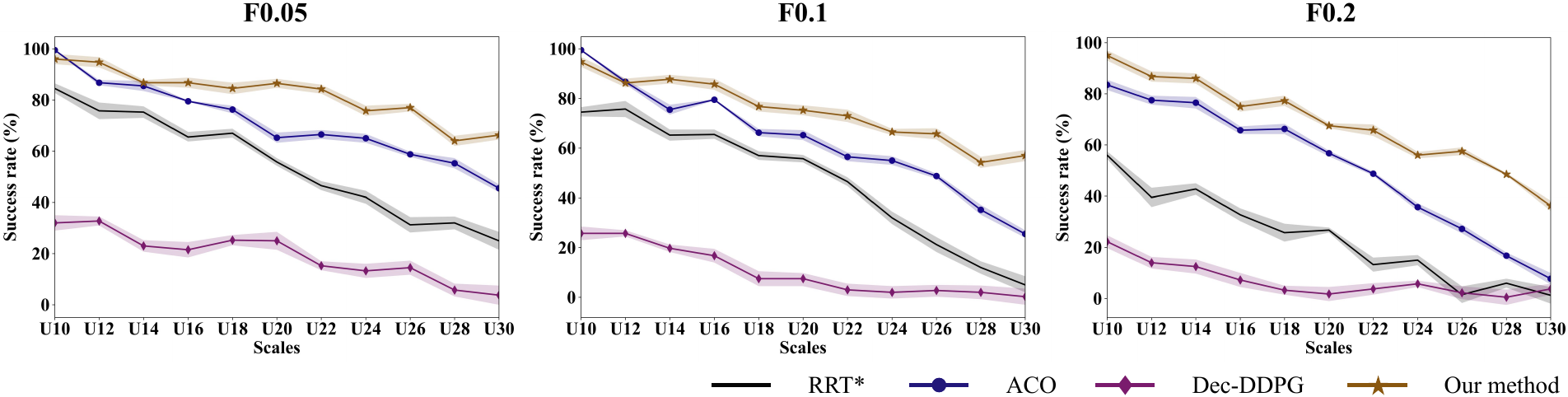}
    \caption{Scalability analysis with varying hazard frequencies. The figures illustrate success rates across scales U10 to U30 for F0.05, F0.1, and F0.2 scenarios, comparing our method against several baselines.}
    \label{fig4}
\end{figure*}

\subsection{Ablation Study}
\label{sec:sample5C}
We incorporate safety guarantees (SG) via a CBF to ensure collision-free paths and provide rapid responsiveness even during sudden replanning. Then, we employ uncertainty mitigation (UM) using a PNN to quantify hazard uncertainty, enhancing the anticipation of potential dangers. Additionally, we adopt observation sharing (OS) via a proximity-weighted approach, enabling effective mutual risk avoidance within the swarm. To evaluate the impact of these components, we conduct ablation studies, with learning curves for the first 30 episodes presented in Fig. \ref{fig3}.

Without SG, the method experiences pronounced training instability and lower episode returns. The absence of CBF allows unsafe actions, leading to frequent collisions, which disrupt learning and cause task failures. Theoretically, CBF constrains the action to ensure forward invariance of the safe set. In scenarios like U9 \& F0.05, the lack of SG results in more significant fluctuations due to increased swarm size, underscoring the critical role of safety guarantees in training.

The method without UM converges more slowly and stabilizes at a suboptimal value, struggling with uncertainty from abrupt hazardous areas. From a theoretical perspective, UM's virtual environment model quantifies uncertainty, minimizing its impact on value function estimation. The learning curves confirm that UM accelerates convergence and achieves higher episode returns compared with the baseline without UM.

Without OS, the method exhibits lower training efficiency and converges to a suboptimal policy due to the lack of integrated neighbors' observations. Theoretically, this inefficiency arises because the agent is operating with only local observation, thereby failing to leverage the collective situational awareness within the swarm. Conversely, OS uses proximity-weighted mean-field interactions to simplify multi-agent interactions into a fixed-size representation, which enables scalable coordination and precise information aggregation. This results in smoother learning curves and faster convergence.

The results demonstrate that these components collectively enable efficient and stable training by ensuring safety, mitigating environmental uncertainty, and utilizing neighbors' observations. Our approach integrates all three components, achieving superior performance, as evidenced by the highest episode returns and smoothest learning curves. These findings validate the critical roles of SG, UM, and OS in multi-UAV pathfinding, particularly in uncertain environments.

\begin{figure*}[t]
    \centering    \includegraphics[width=0.99\textwidth]{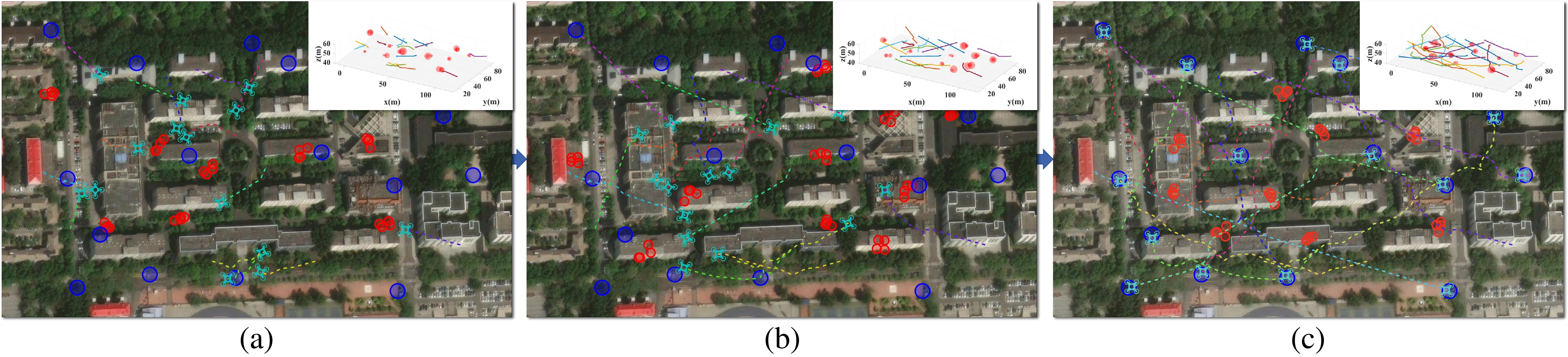}
    \caption{Testing instance of pathfinding in the U12 \& F0.1 scenario.}
    \label{fig_plot}
\end{figure*}

\subsection{Generalization to Larger-Size Instances}
\label{sec:sample5D}
In large-scale multi-agent pathfinding, strategy retraining imposes substantial computational overhead. Thus, we require our trained models to exhibit generalization. The decentralized policies in our method allow direct application to additional agents without retraining. To evaluate the method's generalization, we apply pathfinding networks trained on U9 to instances with 10 to 30 UAVs.
Generalization results are shown in Fig. \ref{fig4}. Since non-learning algorithms lack generalization, they are re-executed for larger scenarios. Our method consistently outperforms RRT*, ACO, and Dec-DDPG across all instances. Notably, in large-scale dynamic scenarios like U30 \& F0.2, our method maintains a 40\% success rate, while baselines drop to 0\%. 
As swarm size increases, pathfinding becomes more challenging, yielding lower returns. However, our method maintains superior solution quality compared with baselines. Specifically, the success rate improvement in U30 is statistically significant with a $p$-value $<$ 0.01, demonstrating robust scalability. These results confirm that policies trained on small-scale instances can plan feasible paths across varying scales, demonstrating our method’s generalization for pathfinding in stochastic environments.

To provide a qualitative example of this generalization, Fig. \ref{fig_plot} illustrates a test case where a policy trained on U9 is directly deployed to the U12 scenario. In this instance, the swarm must navigate an environment with 10 abrupt hazardous areas changing at a frequency of 0.1 Hz. The sequence of snapshots, from (a) to (c), visualizes the collective pathfinding process over time. Despite the increased number of agents and environmental complexity, the UAVs successfully compute and execute collision-free paths, demonstrating favorable avoidance maneuvers. This result confirms that the learned policies can generalize to more crowded and complex scenarios without retraining.

\begin{figure*}[t]
    \centering    \includegraphics[width=0.99\textwidth]{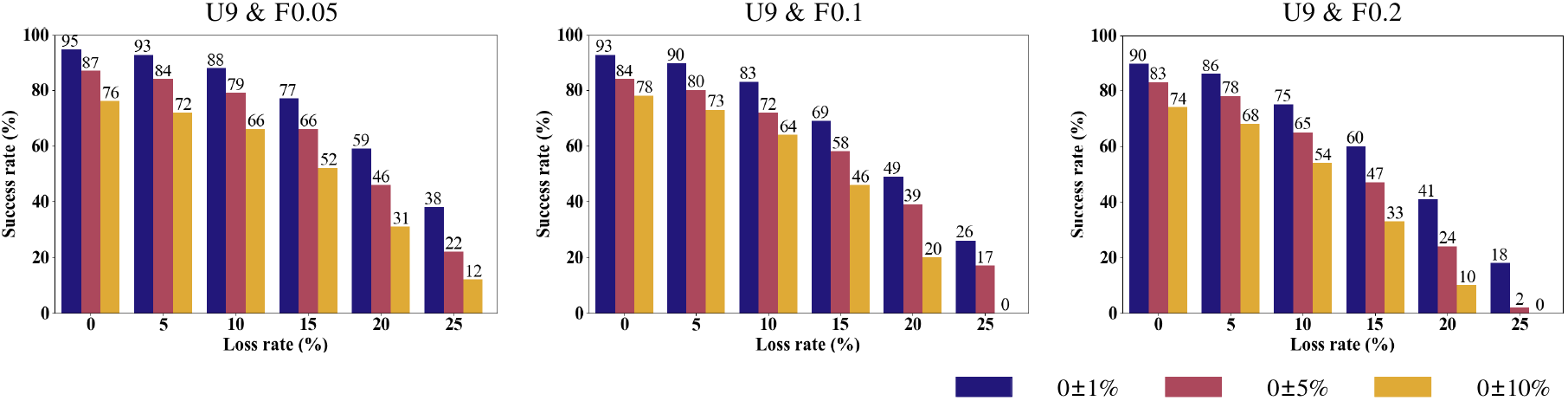}
    \caption{Robustness to communication uncertainty. The figures plot success rate versus message loss rate, with the legend denoting different levels of Gaussian sensor noise.}
    \label{fig5}
\end{figure*}

\begin{figure*}[t]
    \centering    \includegraphics[width=0.99\textwidth]{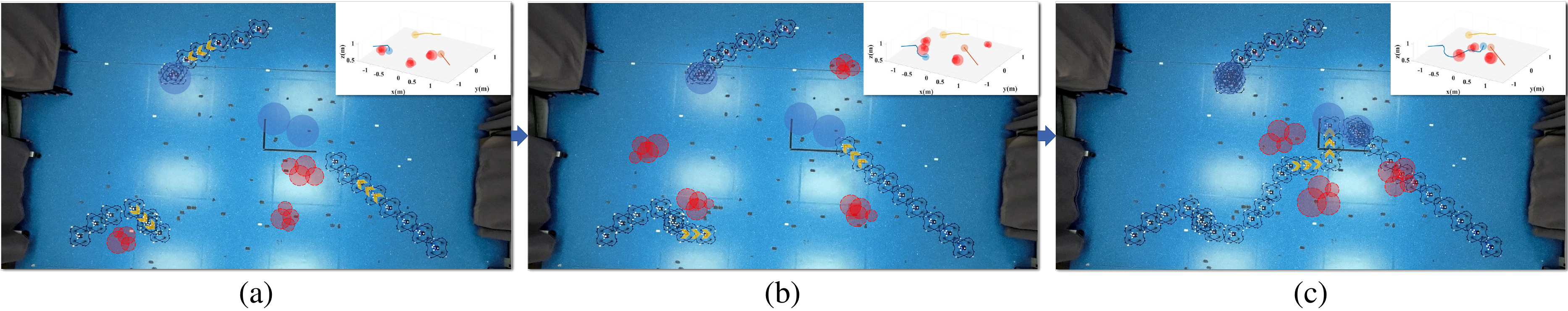}
    \caption{Pathfinding process for UAVs in the indoor real-robot system.}
    \label{fig_indoor}
\end{figure*}

\begin{figure*}[t]
    \centering    \includegraphics[width=0.99\textwidth]{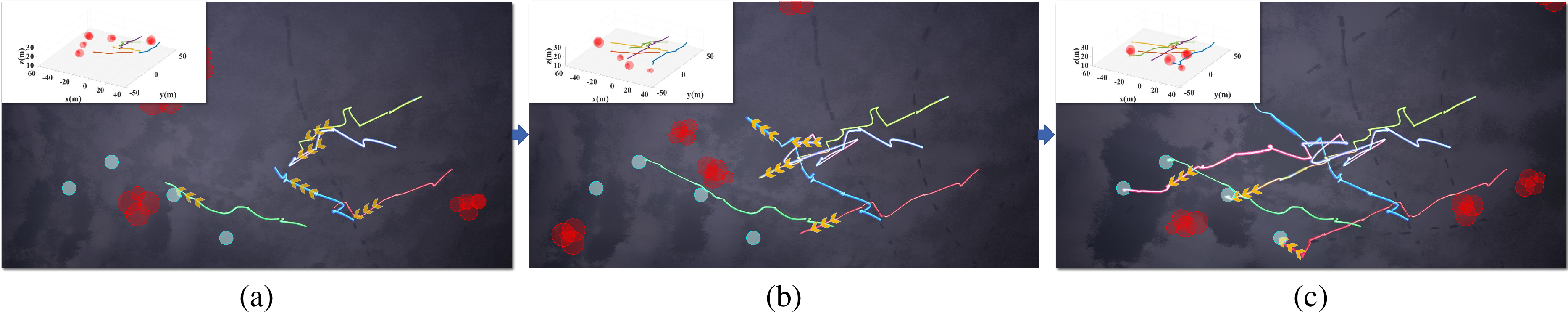}
    \caption{Pathfinding process for UAVs in the outdoor real-robot system.}
    \label{fig_outdoor}
\end{figure*}

\subsection{Deployment in Real-Robot System}
\label{sec:sample5E}

We first evaluate the framework's robustness against communication uncertainties to validate the practical feasibility of our communication-based observation sharing mechanism. We conduct simulation experiments to analyze the impact of inter-agent message loss and sensor noise on mission success, before proceeding to the physical hardware experiments. The results are shown in Fig. \ref{fig5}, which provide an assessment of the method in instances involving non-ideal conditions.
We deploy the trained policy into scenarios with varying hazard change frequencies, where the observations and communications are subjected to different loss rates and noise levels. Although the success rate decreases as the uncertainty increases, it remains above 72\% when the message loss rate is below 10\% and the sensor noise error is within ±5\% in the U9 \& F0.1 scenario. These results confirm that our method is robust enough for practical deployment, effectively handling the imperfections inherent in real-world wireless channels.

For the indoor experiments, we focus on verifying the core collision avoidance and pathfinding capabilities in a 3 m$\times$ 3 m$\times$ 1.5 m setting. The setup, shown in Fig. \ref{fig4_0}(b), utilizes three Crazyflie 2.1 quadcopters, each with a nominal speed of 0.3 m/s and a collision radius of 0.05 m. Their positions are tracked by a Nokov motion capture system at 200 Hz, which provides high-fidelity localization with sub-millimeter accuracy. A central workstation running ROS 2 dynamically generates 3 to 4 virtual hazardous areas at a frequency of 0.25 Hz by specifying point sets, which are then modeled as collections of overlapping spheres with radii ranging from 0.1 m to 0.2 m. Although the hazards are generated centrally, their information is broadcast to each UAV only when the hazard enters the UAV's local sensing range, strictly adhering to the decentralized execution framework. As shown in Fig. \ref{fig_indoor}, the UAVs responsively adjust their trajectories to maintain safe distances, adhering to the enforced safety buffer of $d_s=$ 0.15 m, exhibiting smooth, collision-free paths while progressing towards their goals.

The outdoor experiments are designed to test the algorithm's adaptability in a larger environment of 100 m$\times$ 100 m$\times$ 20 m. The swarm, depicted in Fig. \ref{fig4_0}(c), consists of five Amovlab Z-410 UAVs, each with a nominal speed of 8 m/s and a collision radius of 0.2 m. Each UAV is equipped with a global positioning system module for global self-localization (1.5--2.5 m accuracy) and an ultra wide band sensor for precise inter-agent distance measurements (10--30 cm accuracy). Similarly, the ground workstation generates 3 to 5 time-varying virtual hazardous areas at a frequency of 0.1 Hz, modeled with sphere radii between 3 m and 5 m. The experiments are conducted under a southeasterly wind of approximately 6 m/s. As shown in Fig. \ref{fig_outdoor}, the swarm successfully re-plans its trajectory and executes avoidance maneuvers, maintaining a safety buffer defined by $d_s=$ 1 m, when confronted with the dynamically appearing hazards. This demonstrates that the learned policies are robust enough to handle environmental disturbances and communication uncertainties, ultimately ensuring all agents reach their goals safely. A long-exposure photo capturing these flight trajectories is shown in Fig. \ref{fig4_0}(d).
 
\section{CONCLUSIONS}
\label{sec:sample6}
The proposed MARL method has facilitated the advancement in multiple UAV pathfinding, offering a feasible solution for navigating stochastic environments with abrupt hazardous areas. The integration of safety guarantees, uncertainty mitigation, and observation sharing has allowed superior performance and generalization, as demonstrated through extensive experiments. Compared with traditional methods and existing RL approaches, our method has achieved higher planning success rates, faster computation time, and better scalability. These results have highlighted the potential of our approach for real-world applications in safety-critical scenarios.

\section{APPENDIX}
\label{sec:sample7}

\subsection{Derivation of the Safety Margin Derivative}
\label{sec:sample7A}
Taking the time derivative of the function $H_{in}^{(t)} \left(\boldsymbol{p}_{u,i}^{(t)}, \boldsymbol{p}_n^{(t)} \right)$ in (\ref{equ24}) yields
\begin{align}\dot{H}_{in}^{(t)} \left(\boldsymbol{p}_{u,i}^{(t)}, \boldsymbol{p}_n^{(t)} \right) = \frac{d}{dt} \left( \|\boldsymbol{p}_{u,i}^{(t)} - \boldsymbol{p}_n^{(t)}\| \right) - \frac{d}{dt} \left( d_{in}^{(t)} + d_s \right).\label{equ_a1}\end{align}
To find the derivative of the norm, let $\boldsymbol{d}(t) = \boldsymbol{ p}_{u,i}^{(t)} - \boldsymbol{p}_n^{(t)}$. We first differentiate its square $\|\boldsymbol{d}(t)\|^2 = \boldsymbol{d}(t)^\top \boldsymbol{d}(t)$, using the vector product rule:
\begin{align}
\frac{d}{dt} \left( \|\boldsymbol{d}(t)\|^2 \right) =& \frac{d}{dt} \left( \boldsymbol{d}(t)^\top \boldsymbol{d}(t) \right) \nonumber \\
=& \dot{\boldsymbol{d}}(t)^\top \boldsymbol{d}(t) + \boldsymbol{d}(t)^\top \dot{\boldsymbol{d}}(t)  \nonumber \\
=& 2 \boldsymbol{d}(t)^\top \dot{\boldsymbol{d}}(t).\label{equ_a2}\end{align}
Applying the scalar chain rule to $\|\boldsymbol{d}(t)\|^2$ also gives
\begin{align}\frac{d}{dt} \left( \|\boldsymbol{d}(t)\|^2 \right) = 2 \|\boldsymbol{d}(t)\| \frac{d}{dt} \left( \|\boldsymbol{d}(t)\| \right).\label{equ_a3}\end{align}
By equating (\ref{equ_a2}) and (\ref{equ_a3}), we find
\begin{align}\frac{d}{dt} \left( \|\boldsymbol{d}(t)\| \right) = \frac{\boldsymbol{d}(t)^\top \dot{\boldsymbol{d}}(t)}{\|\boldsymbol{d}(t)\|}.\label{equ_a4}\end{align}
Finally, substituting $\boldsymbol{d}(t) = \boldsymbol{p}_{u,i}^{(t)} - \boldsymbol{p}_n^{(t)}$ and its derivative $\dot{\boldsymbol{d}}(t) = \boldsymbol{v}_{u,i}^{(t)} - \boldsymbol{v}_n^{(t)}$ into (\ref{equ_a4}) yields the expression for $\dot{H}_{in}$:
\begin{align}\dot{H}_{in}^{(t)} \left(\boldsymbol{p}_{u,i}^{(t)}, \boldsymbol{p}_n^{(t)} \right) =&\frac{(\boldsymbol{p}_{u,i}^{(t)} - \boldsymbol{p}_n^{(t)})^\top (\boldsymbol{v}_{u,i}^{(t)} - \boldsymbol{v}_n^{(t)})}{\|\boldsymbol{p}_{u,i}^{(t)} - \boldsymbol{p}_n^{(t)}\|} + \frac{d}{dt} \left( d_{in}^{(t)} \right)
\nonumber \\
\approx & \frac{(\boldsymbol{p}_{u,i}^{(t)} - \boldsymbol{p}_n^{(t)})^\top (\boldsymbol{v}_{u,i}^{(t)} - \boldsymbol{v}_n^{(t)})}{\|\boldsymbol{p}_{u,i}^{(t)} - \boldsymbol{p}_n^{(t)}\|}.\label{equ_a5}\end{align}
This approximation treats the time derivative of the threshold $\dot{d}_{in}^{(t)}$ as negligible, a simplification justified by the fact that the hazard expansion rate ($\mu_{j,m}=$ 0.1 m/s) is significantly smaller than the UAVs' nominal speed ($v_{nom} =$ 3 m/s).

\subsection{Derivation of CBF Decay Rate Bounds}
\label{sec:sample7A}
From the definition of the safety margin function:
\begin{equation}
H_{in}^{[k]} \left(\boldsymbol{p}_{u,i}^{[k]}, \boldsymbol{p}_n^{[k]} \right) = \|\boldsymbol{p}_{u,i}^{[k]} - \boldsymbol{p}_n^{[k]}\| - (d_{in}^{[k]} + d_t),
\label{equ_ap_24}
\end{equation}
its time derivative is given by
\begin{align}
\dot{H}_{in}^{[k]} \left(\boldsymbol{p}_{u,i}^{[k]}, \boldsymbol{p}_n^{[k]}\right) = \frac{(\boldsymbol{p}_{u,i}^{[k]} - \boldsymbol{p}_n^{[k]})^\top (\boldsymbol{v}_{u,i}^{[k]} - \boldsymbol{v}_n^{[k]})}{\|\boldsymbol{p}_{u,i}^{[k]} - \boldsymbol{p}_n^{[k]}\|}.
\label{equ_ap_25}
\end{align}
Thus, for any neighboring agent, a pointwise bound on the time derivative is
\begin{align}
\dot{H}_{in}^{[k]} \left(\boldsymbol{p}_{u,i}^{[k]}, \boldsymbol{p}_n^{[k]} \right) \geqslant - \|\boldsymbol{v}_{u,i}^{[k]} - \boldsymbol{v}_n^{[k]}\|.
\label{equ_ap_26}
\end{align}
Assuming the relative speed is uniformly bounded by $\|\boldsymbol{v}_{u,i}^{[k]} - \boldsymbol{v}_n^{[k]}\| \leqslant V_{\max}$ and the safety margin maintains a positive lower bound $H_{in}^{[k]} \left(\boldsymbol{p}_{u,i}^{[k]}, \boldsymbol{p}_n^{[k]} \right) \geqslant H_{\text{min}} > 0$ during operation, it follows from (\ref{equ_ap_26}) that a sufficient condition for the CBF inequality $\dot{H}_{in}^{[k]} \left(\boldsymbol{p}_{u,i}^{[k]}, \boldsymbol{p}_n^{[k]}\right) \geqslant -\alpha H_{in}^{[k]} \left(\boldsymbol{p}_{u,i}^{[k]}, \boldsymbol{p}_n^{[k]} \right)$ is $-V_{\max} \geqslant -\alpha H_{\text{min}}$. Consequently, $\alpha$ must satisfy
\begin{align}
\alpha \geqslant \alpha_{\min} := \frac{V_{\text{max}}}{H_{\text{min}}},
\label{equ_ap_27}
\end{align}
to ensure robustness against the worst-case relative motion.
For discrete-time forward invariance, we must also consider discretization errors. Using a second-order expansion, the safety margin at the next time step is approximated as
\begin{equation}
H_{in}^{[k+1]} = H_{in}^{[k]} + \dot{H}_{in}^{[k]}\tau + \frac{1}{2}\ddot{H}_{in}^{[k]}\tau,
\label{equ_ap_28}
\end{equation}
where, for brevity, we omit the inputs of $H_{in}^{[k]}, \dot{H}_{in}^{[k]}$, and $\ddot{H}_{in}^{[k]}$. Define $B:=\sup_k \|\ddot H_{in}^{[k]}\|,$ where $\sup$ denotes the supremum. Combining this with the CBF condition $\dot{H}_{in}^{[k]} \geqslant \alpha H_{in}^{[k]}$, we obtain the one-step lower bound:
\begin{equation}
H_{in}^{[k+1]} \geqslant H_{in}^{[k]}(1-\alpha \tau) - \frac{B}{2} \tau^2.
\label{equ_ap_29}
\end{equation}
To ensure the safety margin remains above $H_{\text{min}}$, a sufficient condition for forward invariance is
\begin{equation}
(1-\alpha \tau)H_{\min} - \frac{B}{2} \tau^2\geqslant 0.
\label{equ_ap_30}
\end{equation}
This implies an upper bound for $\alpha$:
\begin{equation}
\alpha \leqslant \alpha_{\max} := \frac{1}{\tau} \left( 1-\frac{B\tau^2}{2H_{\min}} \right),
\label{equ_ap_31}
\end{equation}
provided that $B \tau^2 < 2 H_{\text{min}}$; otherwise, no positive $\alpha$ satisfies the inequality. Combining \eqref{equ_ap_27} and \eqref{equ_ap_31}, a nonempty admissible interval for $\alpha$ exists if and only if
\begin{equation}
\frac{V_{\text{max}}}{H_{\text{min}}} \leqslant \frac{1}{\tau} \left( 1-\frac{B\tau^2}{2H_{\min}} \right).
\label{equ_ap_32}
\end{equation}
This inequality provides a concrete, testable condition relating the sampling time $\tau$, the worst-case dynamics ($V_{\max}$, $B$), and the required safety margin $H_{\text{min}}$. In practice, $V_{\max}$ and $B$ are estimated from logged trajectories, and $\alpha$ is selected within the resulting interval, typically at the midpoint to ensure robustness. If the interval is empty, either the sampling time $\tau$ must be reduced or the safety margin $H_{\text{min}}$ must be increased.

\subsection{Proof of the Mean-Field Policy Approximation}
\label{sec:sample7B}
In the following, we elaborate on the proof of approximation in (\ref{equ38}).
We decompose the neighbor observation $\boldsymbol{z}_j^{[k]}$ (for $ j \in \mathcal{G}_i$) into the proximity-weighted mean $\boldsymbol{\bar{z}}_i^{[k]}$ and a small deviation $\boldsymbol{\delta z}_{ij}^{[k]}$, such that
\begin{align}
\boldsymbol{z}_j^{[k]}=\boldsymbol{\bar{z}}_i^{[k]}+\boldsymbol{\delta z}_{ij}^{[k]}.
        \label{equ36}
\end{align}
According to (\ref{equ36}), the expression (\ref{equ38}) can be rewritten as
\begin{align}
    \pi_i\left( \boldsymbol{z}_{i} ^{[k]}\right)=\sum_{j\in \mathcal{G}_i}{w_{ij}^{[k]}\pi_i\left( \boldsymbol{z}_i^{[k]},\boldsymbol{\bar{z}}_i^{[k]}+\boldsymbol{\delta z}_{ij}^{[k]} \right)}.
\label{equ39}
\end{align}
Letting  $\pi_i\left( \boldsymbol{z}_i^{[k]},\boldsymbol{\bar{z}}_i^{[k]} \right)=\pi_0$, we perform a second-order Taylor expansion of $\pi_i\left( \boldsymbol{z}_i^{[k]},\boldsymbol{\bar{z}}_i^{[k]}+\boldsymbol{\delta z}_{ij}^{[k]} \right)$ at $\left( \boldsymbol{z}_i^{[k]}, \boldsymbol{\bar{z}}_i^{[k]}\right)$:
\begin{align}
    \pi_i\left( \boldsymbol{z}_{i}^{[k]}\right)= \sum_{j\in \mathcal{G}_i} \big( &w_{ij}^{[k]} \pi_0 + w_{ij}^{[k]}\boldsymbol{\delta z}_{ij}^{[k]}\nabla_{\boldsymbol{\bar{z}}_i}^1 \pi_0  \nonumber \\
    & +\frac{1}{2} w_{ij}^{[k]}(\boldsymbol{\delta z}_{ij}^{[k]})^2 \nabla _{\boldsymbol{\bar{z}}_i}^2\pi_0 \big),
\label{equ40}
\end{align}
where $\nabla_{x}^n F\left( x\right)$ represents the $n$-th partial derivative of $F\left( x\right)$ with respect to $x$. Substituting the summation into each term and expanding it term by term:
\begin{align}
    \pi_i\left( \boldsymbol{z}_{i}^{[k]}\right)=\pi_0 \sum_{j\in \mathcal{G}_i}{w_{ij}^{[k]}} &+ \nabla _{\boldsymbol{\bar{z}}_i}^1 \pi_0 \sum_{j\in \mathcal{G}_i}{w_{ij}^{[k]} \boldsymbol{\delta z}_{ij}^{[k]}} \nonumber \\
    &+\frac{1}{2}\nabla _{\boldsymbol{\bar{z}}_i}^2\pi_0\sum_{j\in \mathcal{G}_i}{w_{ij}^{[k]} (\boldsymbol{\delta z}_{ij}^{[k]})^2}.
\label{equ41}
\end{align}
Substituting into the definition of $\boldsymbol{\bar{z}}_i^{[k]}$ in (\ref{equ32}), we have
\begin{align}
        \displaystyle \sum_{j\in \mathcal{G}_i}{w_{ij}^{[k]}\boldsymbol{\delta z}_{ij}^{[k]}}=&\sum_{j\in \mathcal{G}_i}{w_{ij}^{[k]}\left( \boldsymbol{z}_j^{[k]}-\boldsymbol{\bar{z}}_i^{[k]} \right)} \nonumber \\
        =&\sum_{j\in \mathcal{G}_i}{w_{ij}^{[k]} \boldsymbol{z}_j^{[k]}-\boldsymbol{\bar{z}}_i^{[k]}} \sum_{j\in \mathcal{G}_i}{w_{ij}^{[k]}} \nonumber \\
        =&\boldsymbol{\bar{z}}_i^{[k]}-\boldsymbol{\bar{z}}_i^{[k]} = 0.
        \label{equ37}
\end{align}
Hence, the second term in (\ref{equ41}) vanishes:
\begin{align}
    \pi_i\left( \boldsymbol{z}_{i}^{[k]}\right)=\pi_0+0+\frac{1}{2} \nabla _{\boldsymbol{\bar{z}}_i}^2\pi_0 \sum_{j\in \mathcal{G}_i}{w_{ij}^{[k]} (\boldsymbol{\delta z}_{ij}^{[k]})^2}.
\label{equ42}
\end{align}
Therefore,
\begin{align}
   \pi_i\left( \boldsymbol{z}_{i}^{[k]}\right) \approx \pi_0=\pi_i\left( \boldsymbol{z}_i^{[k]},\boldsymbol{\bar{z}}_i^{[k]} \right).
\label{equ43}
\end{align}
The approximation holds with a second-order small error, which guarantees that the policy learned using these approximations converges to a near-optimal policy.

\subsection{Description of Resultant Acceleration for UAVs}
\label{sec:sample7C}
\begin{figure}[t]
    \centering    \includegraphics[width=0.36\textwidth]{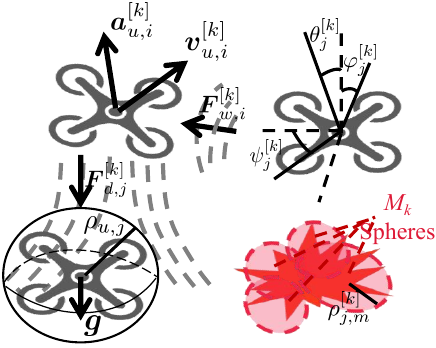}
    \caption{Diagram of the UAVs and hazardous areas.}
    \label{fig_uav}
\end{figure}

In our simulation platform, the resultant acceleration is $ \boldsymbol{a}_{u,i}^{[k]} = \boldsymbol{a}_{r,i}^{[k]} + \boldsymbol{a}_{e,i}^{[k]} + \boldsymbol{g} $, comprising the thrust acceleration $ \boldsymbol{a}_{r,i}^{[k]} $, the external aerodynamic acceleration $ \boldsymbol{a}_{e,i}^{[k]} $, and the gravitational acceleration $\boldsymbol{g} = (0, 0, -g)$, where $g$ is the gravitational constant. The thrust acceleration is bound by $ \|\boldsymbol{a}_{r,i}^{[k]} \| \leqslant a_{\max}$, where $a_{\max}$ is the maximum allowable thrust acceleration. Based on \cite{chen2021adaptive,cabuk2024modeling}, the external acceleration is given by
\begin{equation}
    \boldsymbol{a}_{e,i}^{[k]} = \frac{1}{2m_i}\left( \boldsymbol F_{w, i}^{[k]} + \boldsymbol e_3 \sum_{j\in\mathcal{G}_i} F_{d, j}^{[k]} \right),
\end{equation}
where $ m_i $ denotes the mass of the $ i $-th UAV.
The propeller downwash force is given by $ F_{d, j}^{[k]} = -12.5 S_{i} T_{j}^{[k]}/(\text{Pi}(\boldsymbol{e}_3^T\boldsymbol{p}_{u,j}^{[k]}-\boldsymbol{e}_3^T\boldsymbol{p}_{u,i}^{[k]})^2)$, where Pi is the mathematical constant ($\approx$ 3.14159), when the $i$-th UAV is positioned below the $j$-th UAV in the $z$-coordinate, and $F_{d,j}^{[k]} = 0$ otherwise. The thrust is $T_{j}^{[k]} = m_j\boldsymbol{e}_3^T(\boldsymbol{a}_{r,j}^{[k]}+\boldsymbol{g})$ and $S_{i}$ is the top surface area of the $i$-th UAV. The wind interference force is given by $\boldsymbol F_{w, i}^{[k]}=-C_w^{[k]}\rho_{a}S_i\|\boldsymbol{v}_{a,i}^{[k]}\|\boldsymbol{v}_{a,i}^{[k]}$, where the drag coefficient is $C_w^{[k]}=2m_ig/(\rho_{a}S_i\|\boldsymbol{v}_{a,i}^{[k]}\|^2)$. $\rho_{a}$ is the air density, and $\boldsymbol{v}_{a,i}^{[k]} = \boldsymbol{v}_{u,i}^{[k]} - \boldsymbol{v}_w^{[k]}$ represents the airspeed, where $\boldsymbol{v}_w^{[k]}$ is the wind velocity. The diagram of the UAVs and hazards is shown in Fig. \ref{fig_uav}.

\subsection{Parameter Classification}
\label{sec:sample7D}
The parameters in our strategy can be divided into three main classes according to their function:

\begin{enumerate}
\item Standard RL hyperparameters, including foundational parameters common to most DDPG-style MARL algorithms, such as the discount factor $\gamma$, actor/critic learning rates, and soft update weight. These parameters are foundational to the learning process.
\item Framework-specific weights, which contain the core tuning parameters of our novel framework, including the uncertainty loss weight $\epsilon_u$ and safety reward weight $\epsilon_t$.
\item Safety-critical parameters, including the CBF's safe distance $d_s$ and decay rate $\alpha$, the reward function's threat distance $d_t$ and collision penalty $r_c$, and the number of observed neighbors $N$.

\end{enumerate}

\subsection{Parameter Impact and Selection}
\label{sec:sample7E}
The standard RL hyperparameters are set to values widely recognized in \cite{chai2024multi,wu2024model} for promoting stable convergence in continuous action spaces. The network architectures, detailed in Table \ref{table:net_arch}, provide sufficient capacity for this complex multi-agent task without being computationally prohibitive.

The framework-specific weights are crucial to the performance of our method. The uncertainty weight $\epsilon_u$ is critical for adapting to dynamic environments, and we employ a linear annealing schedule from 0.8 down to 0.2. This strategy compels the agent to first learn a robust policy that avoids uncertainty before gradually shifting its focus to finding more efficient routes. The safety reward weight $\epsilon_t$, set to 1, balances the agent's imperative to avoid the threat zone against its goal-seeking objective, and is tuned to provide a good balance between the normalized goal and safety penalty terms.

The safety-critical parameters are selected based on theoretical and practical considerations. It is crucial to distinguish between the two safety distances: $d_s$ and $d_t$. The safe distance $d_s$, set to 1 m, defines the hard safety boundary used by the CBF during execution for enforcement. In contrast, the threat distance for reward $d_t$, set to a larger value of 4 m, defines a soft "warning zone" used only during training to shape the reward signal and teach the agent to maintain a cautious buffer. The collision penalty $r_c$ is set to -500 to provide a strong learning signal against catastrophic failures. The selection of the CBF decay rate $\alpha$ is guided by the theoretical analysis in Appendix-B, chosen from within the valid interval $\left[\alpha_{\min}, \alpha_{\max}\right]$. The number of observed neighbors $N$ is set to 3, as this value is experimentally found to be a robust compromise between situational awareness and system efficiency. A smaller number would limit the agent's perception of the local swarm density, increasing the risk of collisions in crowded scenarios. Conversely, a larger number would increase both computational and communication overhead with diminishing returns on performance improvement.

\bibliographystyle{IEEEtran}
\bibliography{icra}

\begin{IEEEbiography}[{\includegraphics[width=1in,height=1.25in,clip,keepaspectratio]{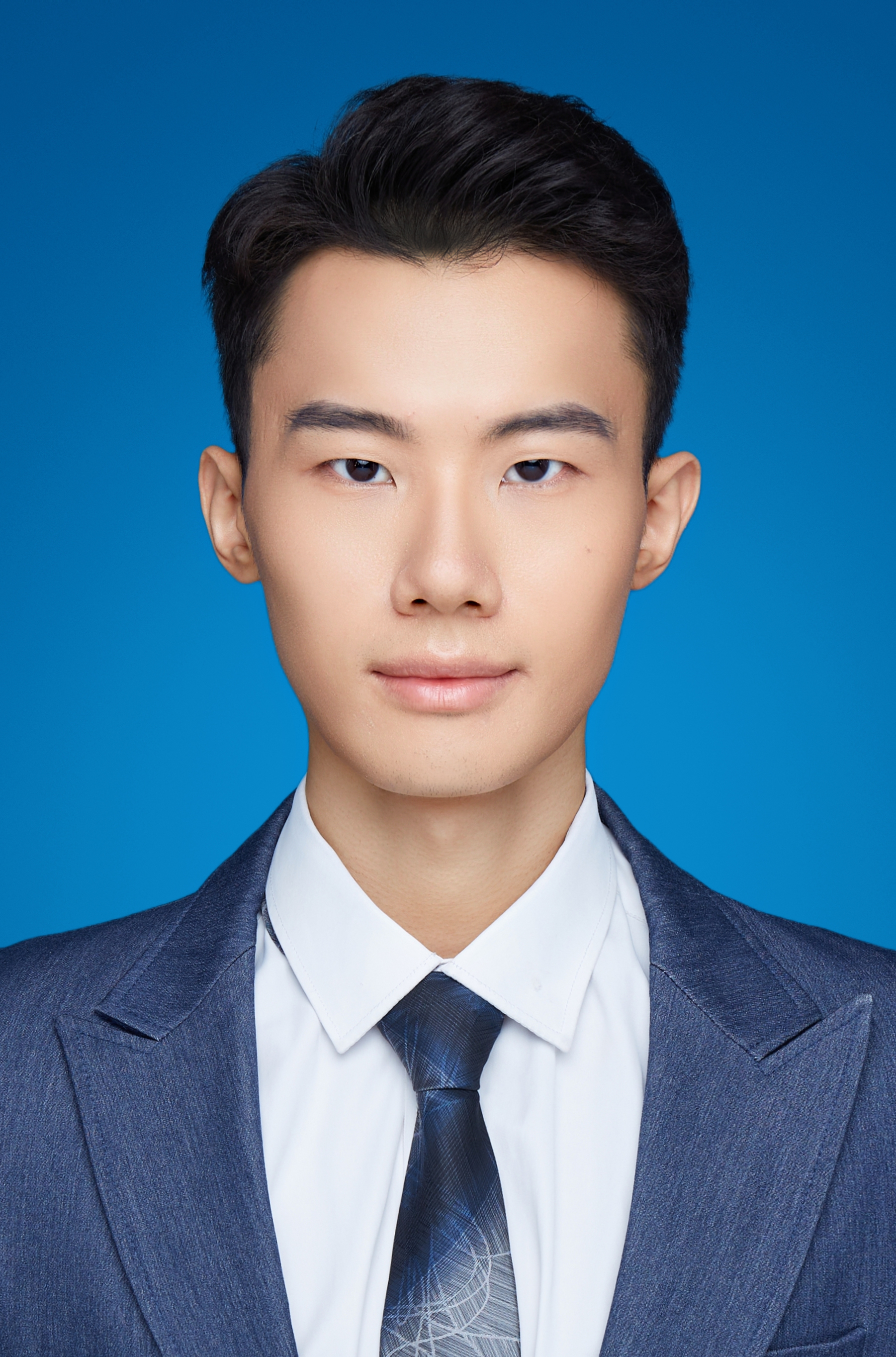}}]{Qizhen Wu} received the B.S. degree in aeronautical and astronautical engineering from Sun Yat--sen University, Guangzhou, China, in 2022. He is currently pursuing the Ph.D. degree with the School of Automation Science and Electrical Engineering, Beihang University, Beijing, China. His current research interests include reinforcement learning, robotic control, and swarm confrontation.
\end{IEEEbiography}
\begin{IEEEbiography}[{\includegraphics[width=1in,height=1.25in,clip,keepaspectratio]{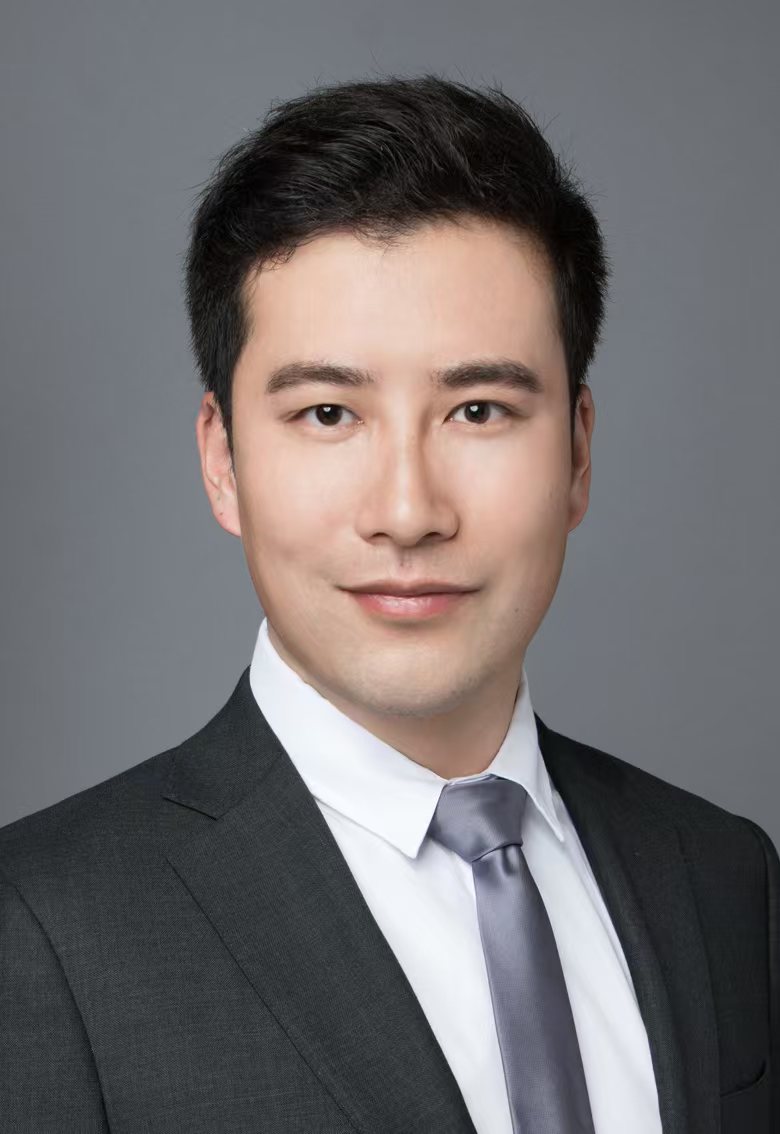}}]{Lei Chen}
received the Ph.D. degree in control
theory and engineering from Southeast University,
Nanjing, China, in 2018.
He was a visiting Ph.D. student with the Royal
Melbourne Institute of Technology University,
Melbourne, VIC, Australia, and Okayama
Prefectural University, Soja, Japan. From 2018
to 2020, he was a Postdoctoral Fellow with
the School of Automation Science and Electrical
Engineering, Beihang University, Beijing, China. He
is currently with the Advanced Research Institute
of Multidisciplinary Science, Beijing Institute of Technology, Beijing, as an
Associate Research Fellow. His current research interests include complex
networks, characteristic model, spacecraft control, and network control.
\end{IEEEbiography}
\begin{IEEEbiography}[{\includegraphics[width=1in,height=1.25in,clip,keepaspectratio]{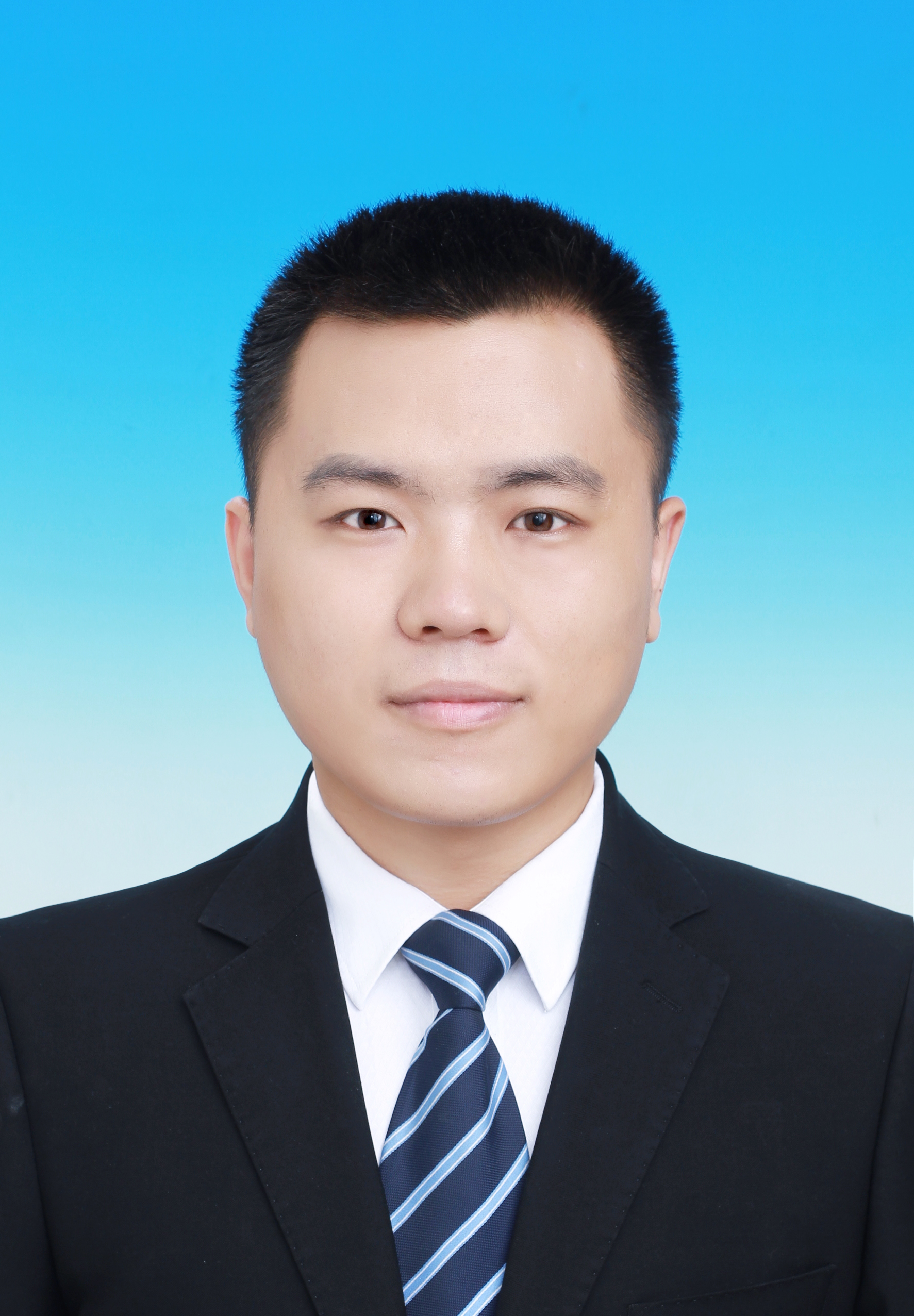}}]{Kexin Liu}
received the M.Sc. degree in control
science and engineering from Shandong University,
Jinan, China, in 2013, and the Ph.D. degree in
system theory from the Academy of Mathematics
and Systems Science, Chinese Academy of Sciences,
Beijing, China, in 2016.
From 2016 to 2018, he was a Postdoctoral
Fellow with Peking University, Beijing. He is currently an Associated Professor with the School
of Automation Science and Electrical Engineering,
Beihang University, Beijing. His current research interests include multi--agent systems and complex networks.
\end{IEEEbiography}
\begin{IEEEbiography}[{\includegraphics[width=1in,height=1.25in,clip,keepaspectratio]{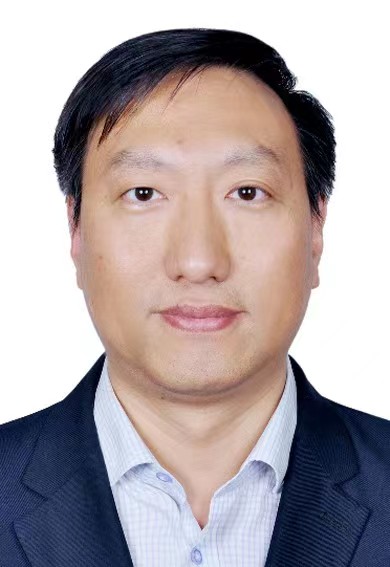}}]{Jinhu L\"u}
(Fellow, IEEE) received the Ph.D.
degree in applied mathematics from the Academy
of Mathematics and Systems Science, Chinese
Academy of Sciences, Beijing, China, in 2002.
He was a Professor with RMIT University,
Melbourne, VIC, Australia, and a Visiting Fellow
with Princeton University, Princeton, NJ, USA.
He is currently the Vice President of Beihang University, Beijing, China. He is also a Professor
with the Academy of Mathematics and Systems
Science, Chinese Academy of Sciences. He is a Chief Scientist of the National
Key Research and Development Program of China and a Leading Scientist
of Innovative Research Groups of the National Natural Science Foundation
of China. His current research interests include complex networks, industrial
Internet, network dynamics, and cooperation control.

Prof. L\"u was a recipient of the Prestigious Ho Leung Ho Lee Foundation
Award in 2015, the National Innovation Competition Award in 2020, the
State Natural Science Award three times from the Chinese Government
in 2008, 2012, and 2016, respectively, the Australian Research Council
Future Fellowships Award in 2009, the National Natural Science Fund for
Distinguished Young Scholars Award, and the Highly Cited Researcher Award
in engineering from 2014 to 2020. He is/was an Editor in various ranks for
15 SCI journals, including the Co--Editor--in--Chief of IEEE TRANSACTIONS
ON INDUSTRIAL INFORMATICS. He served as a member in the Fellows
Evaluating Committee of the IEEE CASS, the IEEE CIS, and the IEEE IES.
He was the General Co--Chair of IECON 2017. He is a Fellow of CAA.
\end{IEEEbiography}

\end{document}